\documentclass{article} 
\usepackage{colm2024_conference}

\usepackage{booktabs}
\usepackage{graphicx}
\usepackage{enumitem}
\usepackage{wrapfig}
\usepackage{algorithm}
\usepackage{algpseudocode}
\usepackage{natbib}
\usepackage{makecell}
\usepackage{booktabs}
\usepackage{array}
\usepackage{amsmath}
\usepackage{amssymb}
\usepackage{amsfonts}
\usepackage{threeparttable}
\usepackage{multirow}
\usepackage{verbatim}
\usepackage{caption}
\usepackage{placeins}
\usepackage{longtable}
\usepackage{ragged2e}      
\usepackage{array}         
\usepackage{supertabular}
\usepackage{CJKutf8}
\usepackage{array} 
\usepackage[utf8]{inputenc} 
\usepackage[T1]{fontenc}
\usepackage[french,vietnamese,mongolian,greek,english]{babel}
\usepackage{textcomp}
\usepackage{pifont}
\usepackage{xcolor}
\usepackage{enumitem}
\usepackage{tablefootnote}
\usepackage{xspace}
\usepackage{textcomp}
\usepackage{makecell}
\usepackage{lscape} 
\usepackage{siunitx}
\usepackage{listings}
\usepackage{xcolor}
\usepackage{colortbl}
\definecolor{kuaishoublue}{HTML}{6D9EEB}
\hypersetup{
    colorlinks=true,   
}
\definecolor{dt}{gray}{0.7}
\usepackage{pifont}       
\usepackage{bbding}       
\usepackage{fontawesome}
\newcolumntype{L}[1]{>{\raggedright\arraybackslash}m{#1}}

\usepackage{scrextend}

\usepackage{tgpagella}
\usepackage{latexsym}
\usepackage[T1]{fontenc}
\usepackage{microtype}
\definecolor{mydarkblue}{rgb}{0,0.08,0.45}
\definecolor{citecolor}{HTML}{0071BC}
\usepackage{url}            
\usepackage{nicefrac}       
\usepackage{changepage}
\usepackage{xargs}          
\usepackage{wrapfig,lipsum,booktabs}
\usepackage{longtable}
\usepackage{subcaption}
\usepackage{endnotes}

\usepackage{fancyvrb}
\usepackage{fvextra}
\usepackage{pgfplots}
\usetikzlibrary{pgfplots.groupplots}
\pgfplotsset{compat=1.3}
\usepackage{tikz}
\usetikzlibrary{patterns}

\usepackage[most]{tcolorbox}
\usepackage[capitalize,noabbrev]{cleveref}
\crefname{section}{Section}{\S\S}
\Crefname{section}{Section}{\S\S}
\crefname{table}{Table}{Tables}
\crefname{figure}{Figure}{Figures}
\crefname{algorithm}{Algorithm}{}
\crefname{equation}{eq.}{}
\crefname{appendix}{Appendix}{}
\crefformat{section}{Section #2#1#3}
\usepackage{multicol}
\usepackage{tcolorbox}

\definecolor{caseaccent}{HTML}{F77F00}
\definecolor{casetitleorange}{HTML}{F77F00}
\definecolor{casenavy}{HTML}{151D35}
\definecolor{casebody}{HTML}{20283D}
\definecolor{casecardbg}{HTML}{FFF8EF}
\definecolor{casecardframe}{HTML}{D9944C}
\newtcolorbox{casecard}[1]{
  breakable,
  enhanced,
  colback=casecardbg,
  colframe=casecardframe,
  boxrule=0.65pt,
  arc=1mm,
  left=7pt,
  right=7pt,
  top=7pt,
  bottom=7pt,
  before skip=0.55em,
  after skip=0.75em,
  borderline north={2.4pt}{0pt}{caseaccent},
  before upper={\noindent{\large\textcolor{casetitleorange}{\bfseries #1}}\par\vspace{0.35em}},
  fontupper=\small\color{casebody}
}
\newenvironment{caseprompt}{\begin{casecard}{Prompt}}{\end{casecard}}
\newenvironment{caseresponse}{\begin{casecard}{Model Response}}{\end{casecard}}
\newcommand{\CaseInputTitle}[1]{{\small\bfseries\textcolor{casenavy}{#1}}}
\usepackage{titlesec}
\titleformat*{\section}{\large\bfseries}

\title{Kwai Keye-VL-2.0 Technical Report}

\author{
\bf Keye Team, Kuaishou Group}

\begin{document}

\maketitle

\begin{abstract}
We introduce Kwai Keye-VL-2.0-30B-A3B, an open-source Mixture-of-Experts (MoE) multimodal foundation model designed to advance long-video understanding and agentic intelligence. To address the challenges of ultra-long contexts, information redundancy, and prohibitive computational costs inherent in hour-level videos, Keye-VL-2.0 is the first to adapt DeepSeek Sparse Attention (DSA) to GQA-based multimodal architectures, enabling lossless 256K context processing while capturing critical frames and long-range temporal dependencies. This architecture is underpinned by a highly optimized training and inference infrastructure, including scalable video I/O, heterogeneous ViT-LM parallelism, and custom DSA kernels that significantly maximize throughput and minimize computational overhead. Furthermore, to overcome the algorithmic dilemma of catastrophic forgetting during multi-task alignment, we introduce Cross-Modal Multi-Teacher On-Policy Distillation (MOPD) paired with Context-RL and Video-RL. By distilling dense token-level teacher feedback from on-policy rollouts back into the MoE backbone—which activates only 3B parameters—Keye-VL-2.0 natively empowers advanced agent collaboration across Code, Tool, and Search scenarios with multimodal self-correction. Extensive evaluations across video understanding, temporal grounding, reasoning, STEM, and agent benchmarks demonstrate that Keye-VL-2.0-30B-A3B achieves state-of-the-art performance among models of similar scale, particularly excelling in fine-grained temporal localization on TimeLens and long-video comprehension on Video-MME-v2 and LongVideoBench. We release our model checkpoints to accelerate community progress toward scalable and robust multimodal agentic applications.

\end{abstract}

\begin{center}
\centering
\includegraphics[width=0.72\textwidth]{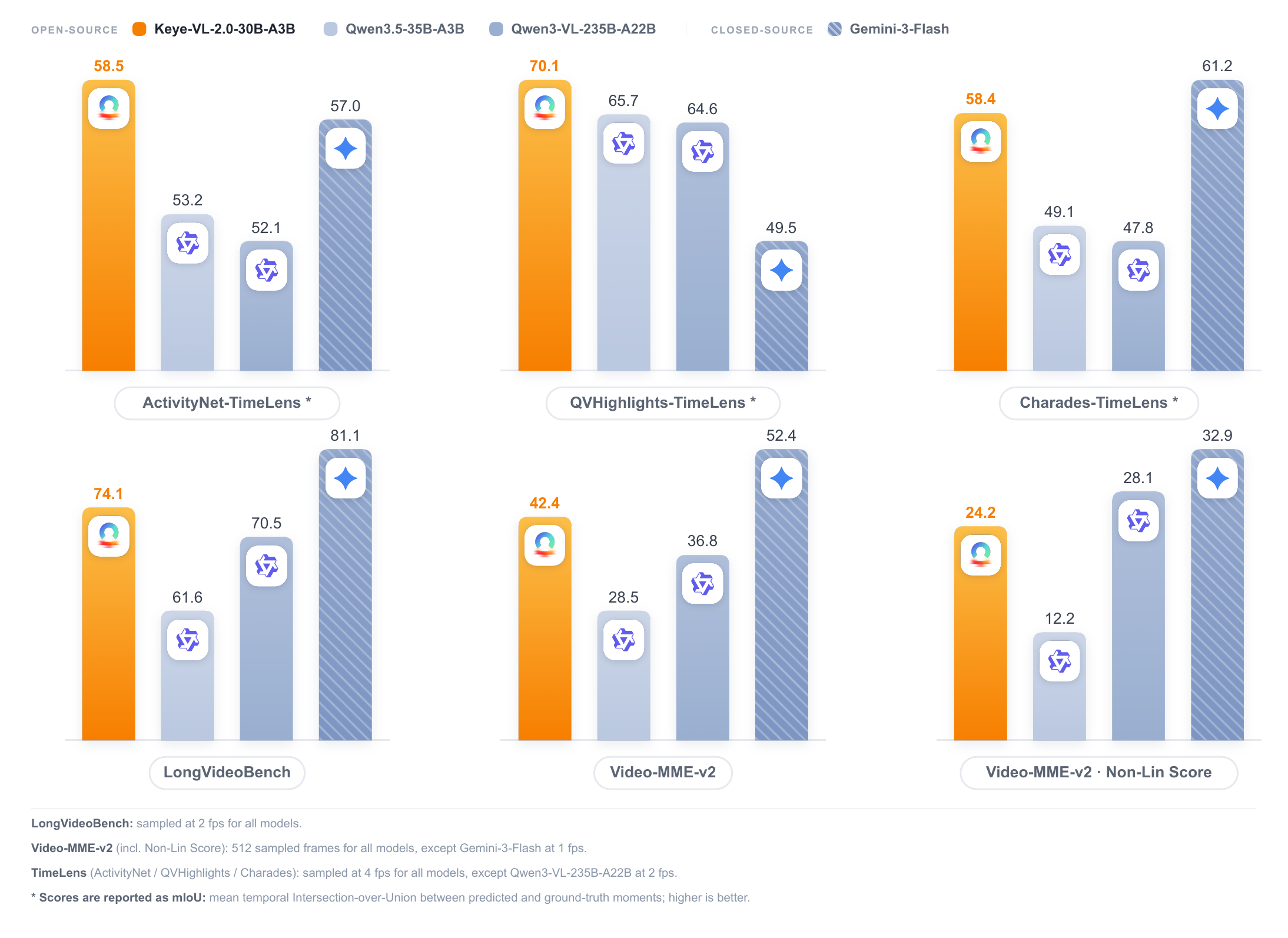}
\captionof{figure}{\textbf{Performance Comparison of Keye-VL-2.0-30B-A3B.} Our model demonstrates leading capabilities against open-source models (e.g., Qwen3.5-35B-A3B, Qwen3-VL-235B-A22B) and closed-source models (Gemini-3-Flash) across fine-grained temporal localization (ActivityNet, QVHighlights, and Charades under the TimeLens framework) and extreme long-video understanding (LongVideoBench, Video-MME-v2).}\label{fig:teaser}
\end{center}



\newpage
{
  \setstretch{0.7}
  \tableofcontents
  \noindent\hrulefill
}
\newpage

\section{Introduction}
In recent years, Large Language Models (LLMs) are rapidly evolving, integrating deeper reasoning and extending into complex multimodal domains. Recent advances, such as OpenAI GPT-5.5 (\cite{openai2026gpt55}), Claude Opus 4.8 (\cite{anthropic2026claude}), Gemini 3.5 Flash (\cite{deepmind2026gemini}), and Qwen3.7 (\cite{qwenteam2026qwen}), demonstrate substantial progress in multimodal reasoning, long-context understanding, and specialized tool execution. These models increasingly exhibit the ability to decompose complex problems and process dynamic, information-dense visual streams. 

Building upon the strong foundation of our previous works, Keye-VL (\cite{kwaikeye2025vl}) and Keye-VL-1.5 (\cite{yang2025kwaikeyevl15}), which established robust vision-language alignment and state-of-the-art short-video comprehension, we introduce Kwai Keye-VL-2.0-30B-A3B. As we push the model's frontier from short-form visual perception to long-horizon agentic reasoning, two critical roadblocks naturally emerge. First is the infrastructural bottleneck: scaling to extreme video contexts incurs prohibitive computational and memory costs. Second is the algorithmic dilemma: integrating complex, heterogeneous agent tasks often induces catastrophic forgetting of foundational reasoning capabilities. In this report, we detail how Keye-VL-2.0 elegantly resolves these dual challenges by introducing paradigm-shifting improvements in architecture and alignment, focusing on the following two key aspects.

\textbf{Extreme Context Scaling via Multimodal DSA.} A key insight from extending multimodal models to 256K contexts is that standard dense attention mechanisms inevitably lead to catastrophic KV cache expansion and computational walls, forcing models to sacrifice temporal continuousness through aggressive frame subsampling. To overcome this latency and scalability limit, Keye-VL-2.0 introduces a 30-billion-parameter Mixture-of-Experts (MoE)(\cite{shazeer2017outrageously}) foundation with only 3 billion active parameters, ensuring exceptional deployment efficiency. Crucially, we pioneer the application of Multimodal DeepSeek Sparse Attention (DSA)(\cite{deepseek2025v32}) in the visual domain. By compressing and sparsifying video feature aggregation, DSA effectively constrains the linear growth of the KV cache. This design allows Keye-VL-2.0 to process 256K extreme long-video contexts losslessly, transforming video understanding from frame-limited perception to global-context reasoning while maintaining high inference efficiency.

\textbf{Resolving Modality Conflict via Cross-Modal MOPD.} Existing models often suffer from the "Multimodal Alignment Dilemma": directly injecting complex video understanding and tool-use capabilities often triggers catastrophic forgetting, degrading the model's foundational STEM, mathematical, and linguistic reasoning abilities. To circumvent the ambiguity and instability of end-to-end co-optimization, we introduce Cross-Modal Multi-Teacher On-Policy Distillation (MOPD)(\cite{mimoteam2026v2flash}). Through an innovative dynamic routing mechanism, MOPD leverages specialized teacher models to provide dense token-level feedback on student-generated trajectories across varying modalities and tasks during post-training. By supervising on-policy rollouts, MOPD effectively isolates task-specific expertise, which is subsequently distilled and seamlessly merged back into the unified MoE backbone. This bidirectional alignment ensures that Keye-VL-2.0 achieves substantial leaps in native agent capabilities (e.g., Code, Tool Use, Web Search) while robustly preserving its general-purpose reasoning baselines. During post-training, we further apply Context-RL(\cite{lu2026contextrl}) and Video-RL, equipped with bucket advantage scaling, to stabilize long-sequence decision trees and systematically reduce visual hallucinations.

As illustrated in Figure \ref{fig:teaser}, Keye-VL-2.0-30B-A3B demonstrates highly competitive performance against both open-source and closed-source state-of-the-art models. It achieves top-tier results in fine-grained temporal localization across TimeLens benchmarks (\cite{zhang2025timelens}) (ActivityNet, QVHighlights, Charades), substantially outperforming leading models such as Gemini-3-Flash~(\cite{deepmind2025gemini3flash}) on several temporal grounding settings. Furthermore, it exhibits a non-linear capability scale-up on extreme long-context evaluations, including Video-MME-v2 (\cite{fu2026videomme2}) and LongVideoBench (\cite{wu2024longvideobench}), proving its robustness in processing extensive temporal information. By resolving the dual challenges of context scaling and multi-task capability conflict, we offer a powerful, efficient, and versatile multimodal foundation for the open-source community, enabling researchers and developers to explore, refine, and deploy scalable multimodal applications.

\section{Model Architecture}

 Following the standard multimodal large language model (MLLM) paradigm, the model consists of four core components:

\begin{itemize}
    \item \textbf{Vision Encoder (ViT):} inherited from Keye-VL-1.5-8B~(\cite{kwaikeye2025vl}) to extract visual features from images and video frames.
    \item \textbf{Language Decoder (LLM):} built on Qwen3-30B-A3B-Thinking-2507~(\cite{qwen3}), providing strong general knowledge, instruction following, and reasoning capabilities.
    \item \textbf{MLP Projector:} randomly initialized and trained in Stage~0 (Section \ref{stage_0}) to align visual features with the LLM representation space.
    \item \textbf{Sparse Attention Module:} a GQA-compatible DSA design that combines global MQA-based indexing with grouped GQA aggregation, enabling efficient long-context multimodal modeling.
\end{itemize}

On top of this backbone, we introduce three architectural designs for high-resolution and long-context multimodal understanding: a native-resolution vision encoder, a unified visual encoding strategy for images and videos, and DSA-based sparse attention for 256K multimodal contexts.


\subsection{Native-Resolution Vision Encoder}

The visual encoders of MLLMs have gradually shifted from reusing fixed-resolution backbones to native-resolution modeling. Conventional fixed-resolution ViTs are usually pre-trained for contrastive image-text matching, and their coarse-grained representations are often insufficient for detail-sensitive downstream tasks such as OCR, fine-grained recognition, document understanding, and video understanding.

Two main approaches have been explored for high-resolution inputs. Dynamic tiling methods, such as InternVL3~(\cite{internvl3}) and MiniCPM-V~(\cite{minicpm_v}), split large images into smaller crops before feeding them into fixed-resolution encoders. While effective, tiling may disrupt global structure and introduce redundant computation. Native-resolution methods, such as NaViT~(\cite{navit}), Qwen3.5~(\cite{qwen3.5}), K2.5~(\cite{team2026kimi}), and MiMO~(\cite{mimoteam2026v2flash}), instead preserve the original image size and aspect ratio. Following this direction, Keye-VL-2.0-30B-A3B encodes images and videos at their native resolutions, avoiding unnecessary cropping or tiling and preserving both global structure and local details.

This design is especially important for document, OCR, chart, and video scenarios, where a small local region may determine the final answer. By keeping the original aspect ratio throughout the visual pipeline, the encoder can preserve layout relations, object geometry, and fine-grained text signals that are easily distorted by fixed-size resizing.

\paragraph{Adaptive Position Encoding.}
The visual encoder inherits the ViT backbone from Keye-VL-1.5, which is based on SigLIP-400M-384-14~(\cite{siglip}). To support variable resolutions while retaining pre-training benefits, we interpolate the fixed absolute learnable position embeddings, allowing them to scale with the input size.

\paragraph{2D RoPE.}
On top of adaptive absolute position encodings, we introduce 2D Rotary Position Embedding (2D RoPE). This improves spatial modeling and extrapolation across visual dimensions, especially for extremely high-resolution images.

\paragraph{Sequence Packing.}
We combine adaptive position encodings and 2D RoPE with NaViT's Patch n' Pack mechanism and FlashAttention. Samples with different sizes and aspect ratios can therefore be packed into a single batch without padding waste, improving training throughput under variable resolutions.

\paragraph{Distribution-Aligned ViT Pre-training.}
During ViT pre-training, we optimize the native-resolution architecture with the SigLIP loss and align it with the SigLIP-400M-384-14 text tower. To reduce the supervision-granularity gap between contrastive pre-training and downstream MLLM tasks, we train the visual encoder on the same data distribution as the downstream MLLM. The pre-training corpus contains 500B tokens from large-scale open-source datasets, including DataComp~(\cite{datacomp}), LAION~(\cite{laion}), CC12M~(\cite{cc12m}), PD12M~(\cite{pd12m}), and COCO~(\cite{coco}), together with high-quality internal data.

Overall, these upgrades allow the visual encoder to inherit the strong representation quality of SigLIP while gaining the resolution flexibility required by downstream multimodal reasoning tasks.

\subsection{Unified Visual Encoding}

To provide the language decoder with detailed visual signals, we use a unified dynamic-resolution encoding strategy for both images and videos.

\begin{itemize}
    \item \textbf{Dynamic-Resolution Image Encoding.}
    Static images are encoded directly by the dynamic-resolution ViT. The number of visual tokens is allocated according to the original pixel size, avoiding lossy resizing or cropping.

    \item \textbf{Dynamic-Resolution Video Encoding.}
    For video inputs with varying frame rates, resolutions, and durations, each sampled frame is treated as an independent high-resolution image and encoded by the same visual encoder. To preserve temporal information, we prepend a natural-language timestamp to the visual tokens of each frame during preprocessing. This explicit timestamp injection helps the LLM perceive temporal order, causality, and absolute timing.

    Compared with designing a separate video-specific encoder, this frame-as-image formulation keeps the visual pathway simple and unified. The timestamp text provides temporal anchors in the LLM's native language space, making temporal localization and cross-frame reasoning easier to learn from instruction and RL data.

    \item \textbf{Adaptive Video Pixel Budget.}
    To balance information density and computation, we allocate a base pixel budget by dividing the global budget evenly across videos. In adaptive mode, the budget of each video is scaled according to duration, with thresholds of $256$\,s, $512$\,s, $1024$\,s, and $2048$\,s corresponding to scaling factors of $0.125$, $0.25$, $0.5$, and $1.0$. Videos longer than $2048$\,s use the full base budget. This compresses short, redundant videos more aggressively while allowing longer videos to retain more visual evidence, keeping the total token cost controllable.
\end{itemize}

\subsection{DSA for Long-Context Multimodal Modeling}

Traditional full attention has quadratic complexity with respect to sequence length, making 256K multimodal contexts difficult to support. To address this bottleneck, we integrate DeepSeek Sparse Attention (DSA)~(\cite{deepseek2025v32}) into the decoder attention pathway. Unlike most existing DSA systems that are adapted from MLA, Keye-VL-2.0-30B-A3B integrates DSA with a GQA-based MLLM backbone, providing a practical route for long-context extension in GQA models.

\subsubsection{MQA-Style Lightning Indexer}

DSA contains a Lightning Indexer and a fine-grained token selection mechanism. For throughput, the indexer follows the MQA key-sharing design and computes a global index score $I_{t,s}$ between the current query token $h_t$ and each preceding token $h_s$:
\begin{equation}
    I_{t,s} = \sum_{j=1}^{H^{I}} w_{t,j}^{I} \cdot \mathrm{ReLU}(q_{t,j}^{I} \cdot k_{s}^{I}).
\end{equation}
Here, $H^I$ is the number of indexer heads, $q_{t,j}^I$ and $w_{t,j}^I$ are derived from $h_t$, and $k_s^I$ is the shared key derived from $h_s$. After obtaining the global scores, the Top-$k$ tokens form the sparse index set:
\begin{equation}
    \Omega_t = \{s \mid I_{t,s} \in \mathrm{Top}\text{-}k(I_{t,:})\}.
\end{equation}

By sharing one key head across all query heads, the indexer substantially reduces both computation and memory traffic. Together with FP8 implementation and the ReLU-based scoring function, the Lightning Indexer remains efficient even when the multimodal sequence contains hundreds of thousands of tokens.

\subsubsection{GQA Sparse Aggregation}

In the GQA backbone, query heads are divided into $G$ groups, and heads within each group share a group-specific KV head. We apply the same sparse index set $\Omega_t$ to all groups. For the $g$-th group, the sparse attention output is:
\begin{equation}
    u_{t,g} = \mathrm{Attn}(h_{t,g}, \{c_{s,g} \mid s \in \Omega_t\}).
\end{equation}
The outputs of all groups are concatenated to form the final attention representation. We set $k=2048$, reducing the core attention complexity from $O(L^2)$ to $O(Lk)$, where $L$ is the sequence length and $k \ll L$.

This global MQA-style indexing + GQA aggregation design preserves the representation structure of the GQA backbone while avoiding dense attention over the full context. As a result, the model can perform long-range spatiotemporal and semantic aggregation with much lower memory and compute cost.

\subsubsection{Dense Warm-up and Sparse Adaptation}

We train DSA with a two-stage strategy. In the dense warm-up phase, the main model keeps dense GQA while most parameters are frozen. The goal is to initialize the indexer so that its global distribution can cover the attention distributions of all GQA groups. For query token $t$, we aggregate dense attention scores within each group, normalize them over visual and text tokens, and obtain the target distribution $p_{t,:,g}$. The warm-up loss is:
\begin{equation}
    \mathcal{L}_{\mathrm{warmup}}^{I} =
    \sum_t \sum_{g=1}^{G}
    \mathbb{D}_{KL}(p_{t,:,g} \parallel \mathrm{Softmax}(I_{t,:})).
\end{equation}
This stage uses approximately 2B multimodal tokens.

In the sparse adaptation phase, all parameters are unfrozen and training switches to sparse mode. The indexer continues to align with the main model's attention distribution, but the KL loss is computed only over the selected Top-$k$ token set
\begin{equation}
S_t = \{s \mid I_{t,s} \in \mathrm{Top}\text{-}k(I_{t,:})\}.
\end{equation}
The target distribution is truncated and renormalized over $S_t$:
\begin{equation}
    \mathcal{L}_{\mathrm{sparse}}^{I} =
    \sum_t \sum_{g=1}^{G}
    \mathbb{D}_{KL}(p_{t,S_t,g} \parallel \mathrm{Softmax}(I_{t,S_t})).
\end{equation}
The main model is optimized with the standard next-token prediction loss. To reduce gradient interference, the indexer input is detached from the computation graph. The final objective is:
\begin{equation}
    \mathcal{L}_{\mathrm{total}} =
    \mathcal{L}_{\mathrm{NTP}} + \lambda \mathcal{L}_{\mathrm{sparse}}^{I}.
\end{equation}

This two-stage training strategy avoids forcing the sparse indexer to learn from scratch under sparse supervision. Dense warm-up first aligns the indexer with the behavior of the original attention model, and sparse adaptation then teaches the full model to rely on dynamically selected evidence. In long-context evaluations, this design preserves the capabilities of the dense model while substantially reducing inference overhead.

\section{Pre-Training}


Keye-VL-2.0-30B-A3B is pre-trained through a four-stage curriculum. Stage~0 trains only the Projector to initialize visual-language mapping. Stage~1 performs full-parameter multimodal pre-training at 32K context length. Stage~2 extends the context to 64K and injects task-oriented capabilities such as OCR, VQA, STEM, GUI, grounding, counting, coding, tool use, and search. Stage~3 further extends the context to 256K, focusing on long videos, multi-page documents, multi-document inputs, and long-range agent trajectories.


\subsection{Stage 0 --- Projector Initialization}
\label{stage_0}

Stage~0 establishes the initial connection between the visual encoder and the language model. The ViT and LLM are frozen, and only the Projector is trained to map visual features from the Keye-VL-1.5 ViT into the LLM representation space. This stage uses image-text caption data for direct semantic alignment and image-text interleaved data to expose the Projector to real document-like contexts where multiple images and text appear alternately.

Because only the Projector is updated, Stage~0 provides a low-risk initialization step before full-parameter multimodal training. It allows the LLM to receive visual features in a compatible representation space without disturbing the already learned capabilities of either the ViT or the language backbone.

\subsection{Stage 1 --- General Multimodal Pre-training}

Stage~1 trains all parameters with a maximum sequence length of 32K on approximately 1T tokens. The goal is to establish stable vision-language alignment, image perception, video understanding, OCR recognition, and general language capability.

The training data consist of image-text captions, interleaved image-text data, interleaved video data, pure-text QA, and OCR data. For video learning, each video is split into 15-second segments, each paired with a caption and organized as an interleaved multimodal sequence. This enables the model to understand temporal continuity, scene transitions, and event evolution.

To improve caption quality from large-scale open-source corpora such as LAION, DataComp, COYO, and CC12M~(\cite{laion,datacomp,coyo,cc12m}), we apply two strategies: \emph{Recaption} regenerates captions directly from the images using an expert captioning model to ensure high quality, while \emph{Remake} builds on the original captions, correcting grammatical and expression errors without altering their semantics. OCR data cover diverse types such as stylized text, handwriting, LaTeX formulas, code, charts, and documents, giving the model an early foundation in text recognition and fundamental perception capabilities.

This diversity of data types—spanning image caption, interleaved image-text, interleaved video-text, pure-text, and OCR data—is designed to prevent the model from over-specializing in any single input format. Specifically, image caption data provide dense semantic alignment between an image and its description; interleaved image-text data expose the model to the organization of images and text within webpages and documents; interleaved video-text data introduce temporal and cross-modal structure; and OCR data establish sensitivity to visual text and structured page layouts from the outset of multimodal training.

\begin{figure*}[t]
\centering
\includegraphics[width=0.95\textwidth]{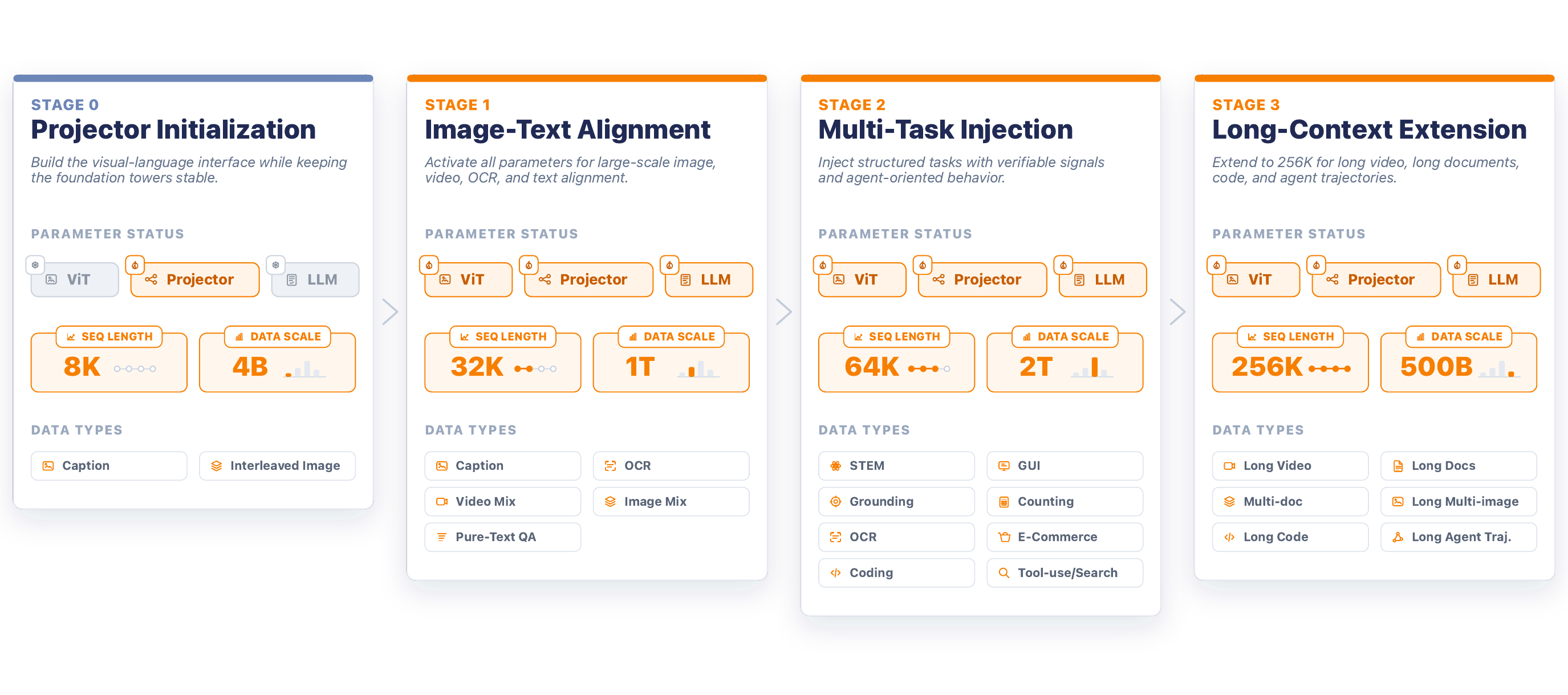}
\caption{\textbf{The Keye-VL-2.0-30B-A3B pre-training pipeline,} following a four-stage curriculum from projector initialization to 256K long-context multimodal training.}
\label{fig:pretraining_stages}
\end{figure*}

\subsection{Stage 2 --- Multi-Task Capability Injection}

Stage~2 continues full-parameter training, extends the context length to 64K, and trains on approximately 2T tokens. Compared with Stage~1, this stage emphasizes task-oriented and verifiable supervision.

\subsubsection{Image-Text Data}

\paragraph{Advanced OCR.}
We further expand the OCR data to receipts and diverse chart types, drawing on both real samples and synthetic samples generated from XML or structured templates. Augmentations such as blur, lighting variation, wrinkles, handwriting variation, and geometric distortion are applied to improve robustness for field extraction, table reading, chart understanding.

Real samples consist of images captured in real-world settings, carrying authentic noise and naturally occurring business layouts, whereas synthetic samples offer precise field-level and structure-level annotations. Combining the two improves both robustness and supervision precision.

\paragraph{Math \& STEM.}
We introduce visual problem-solving examples spanning mathematics, physics, and chemistry, with LLM-based verification to ensure the quality of questions and answers and to remove low-quality samples. This helps the model progress from recognizing formulas and diagrams to reasoning over them. The covered scenarios include geometry diagrams, function plots, experimental apparatus, formulas, and scientific charts. The goal is to enhance the model's generalization across diverse visual scenarios.

\paragraph{Caption Refinement.}
High-quality caption data is retained to preserve vision-language alignment. An 8B caption expert model trained with VCap~(\cite{vcap}) generates detailed descriptions emphasizing attributes, spatial relations, subject binding, and long-description consistency—yielding captions that are more comprehensive, ine-grained, and of higher quality than those used in Stage 1, surpassing even those produced by larger models.

\paragraph{GUI.}
GUI tasks involve screen captures, control metadata, and interaction semantics, covering element localization, control recognition, page understanding, operation description, and GUI-QA. GUI inputs differ substantially from natural images, featuring dense small text, icons, regular layouts, and frequent state changes. Training on dedicated GUI tasks therefore lays the foundation for subsequent vision-based clicking, navigation, and task execution.

\paragraph{Grounding \& Counting.}
For localization and counting, we synthesize instance-pasting data. Candidate objects from COCO~(\cite{coco}) and OpenImages~(\cite{openimages}) are verified by an MLLM and pasted onto backgrounds with controllable quantities and positions, producing precise bounding boxes and count annotations.

\paragraph{General QA, E-Commerce, and Chinese Expansion.}
We supplement general VQA data for object recognition, attributes, scenes, and relations, and introduce Kuaishou e-commerce data for product understanding. To improve Chinese coverage, high-quality English task data are translated into Chinese, regenerated under original answer constraints, and filtered for consistency.

\subsubsection{Pure-Text Data}

Stage~2 retains pure-text training to preserve the LLM's language, reasoning, coding, tool-use, and instruction-following capabilities. The mixture spans pure-text math and STEM reasoning, code corpora for bug fixing and competitive programming, Hermes-style tool-use trajectories~(\cite{hermes}), and search/RAG examples covering single-document QA, multi-document summarization, evidence localization, and multi-hop reasoning over Wiki-based knowledge graphs.

The pure-text component mitigates the degradation of the base model's language-side capabilities during multimodal fine-tuning, while supplying transferable priors for reasoning formats, code generation, tool invocation, and evidence-based answer synthesis.

\subsection{Stage 3 --- Long-Context Extension}

Stage~3 extends the maximum sequence length to 256K while continuing full-parameter training. Long-context and short-context samples are mixed at a ratio of $1{:}1$ to improve ultra-long sequence modeling while preserving performance on conventional inputs.

The data cover long videos, long documents, multi-document inputs, long multi-image conversations, long code contexts, and long-range agent trajectories. The objective is not only to enlarge the context window, but also to improve retrieval, aggregation, and cross-position reasoning over ultra-long multimodal inputs.

For long videos, the model must track key events and integrate evidence across many frame segments. For long documents and multi-document inputs, it must process multi-page OCR, tables, layouts, and long textual evidence. For agent trajectories, it must maintain task state across many tool calls and interaction turns.

\subsection{Data Cleaning and Production Pipeline}

We build a unified data cleaning, deduplication, and production pipeline. Cleaning removes low-quality text, invalid formats, unsafe content, duplicated samples, and unreliable image-text or video-text pairs. Deduplication uses a joint Hash + CLIP strategy: Hash identifies exact or near duplicates, while CLIP similarity detects semantically similar cross-modal duplicates.

For production efficiency, the pipeline supports caption generation, OCR construction, QA generation, translation, quality evaluation, one-click deployment, task monitoring, and checkpoint-based resumption. A dual-queue asynchronous mechanism decouples CPU-side preprocessing from GPU-side inference, improving production throughput by 3--5$\times$ in practice.

\begin{table}[h]
\centering
\caption{Video training setup across pre-training stages.}
\label{tab:video_training_setup}
\small
\setlength{\tabcolsep}{5pt}
\begin{tabular}{
>{\centering\arraybackslash}m{2.5cm}
>{\centering\arraybackslash}m{3.5cm}
>{\centering\arraybackslash}m{2.5cm}
}
\toprule
\textbf{Stage} & \textbf{Max Video Duration} & \textbf{Video Tokens} \\
\midrule
1 & 15s & 24K \\
2 & 15min & 64K \\
3 & 2h & 180K \\
\bottomrule
\end{tabular}
\end{table}

\subsection{Video Pre-Training Curriculum}

To scale from short-video understanding to high-resolution long-video reasoning, we adopt a multi-stage video curriculum, summarized in \Cref{tab:video_training_setup}.

Stage~1 uses short 15-second clips for video-language alignment. Stage~2 increases duration and resolution, and introduces temporal video grounding (TVG) data. Stage~3 extends the maximum duration to 2 hours, requiring the model to identify sparse but critical moments and aggregate long-range evidence. In mid-training, we keep the Stage~3 setup while adding richer tasks such as video captioning, temporal QA, entity recognition, scene understanding, causal reasoning, event ordering, and video counting.

\begin{figure*}[t]
\centering
\includegraphics[width=\textwidth]{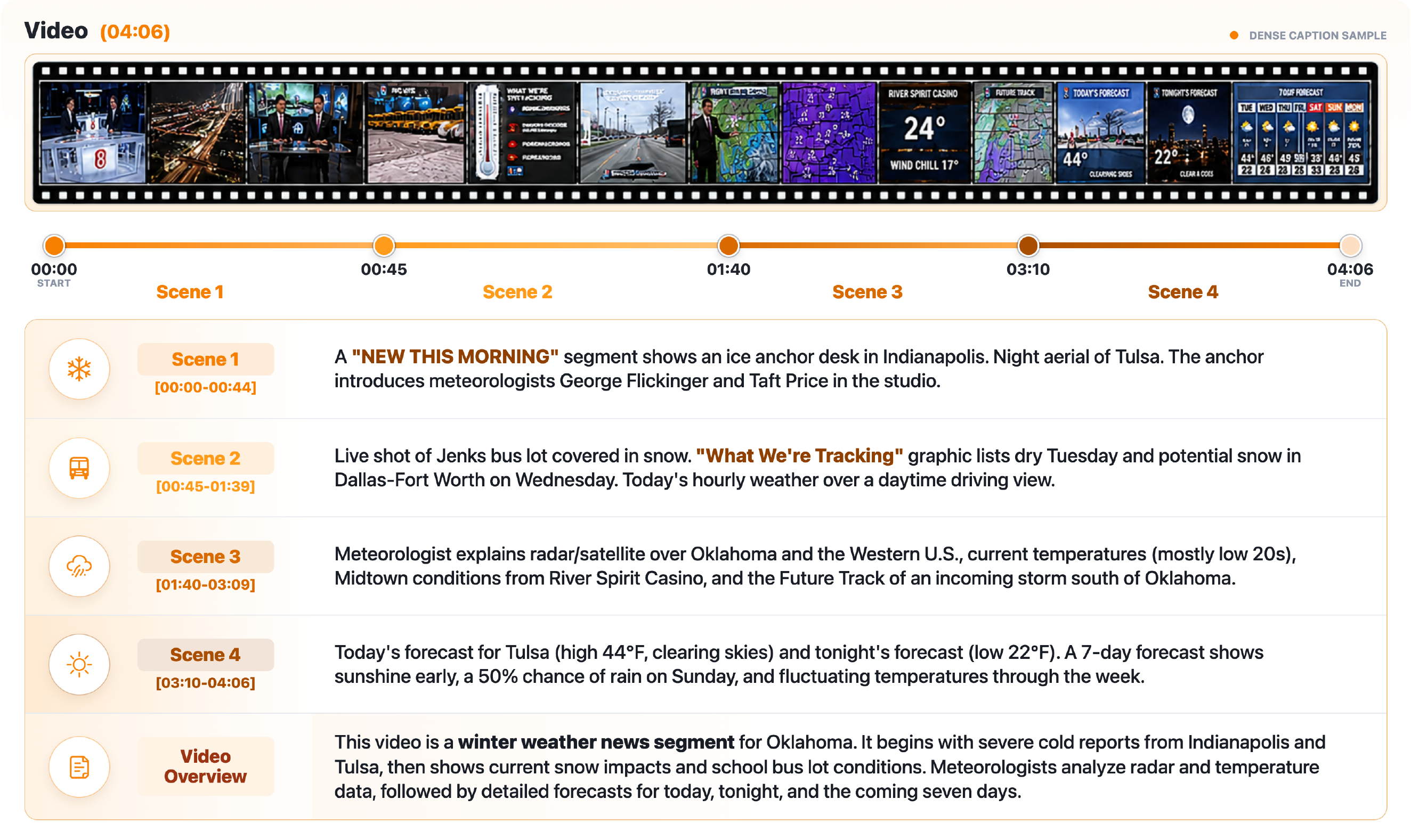}
\caption{\textbf{An example of scene-wise dense caption.} Each video is decomposed into scenes annotated with timestamps, dense captions, and a global overview.}
\label{fig:dense_cap}
\end{figure*}

\paragraph{Scene-Wise Dense Caption.}
We reformulate dense video captioning as structured scene-wise descriptions with start and end timestamps. As shown in Figure~\ref{fig:dense_cap}, this formulation enhances scene-boundary perception and temporal alignment between descriptions and intervals. An expert model is trained to improve annotation efficiency.

\paragraph{Diverse TVG Data.}
Inspired by ETBench~(\cite{etbench}), we construct TVG data covering Referred Action Recognition, Video Highlight Detection, Extractive Video Summarization, and Temporal Event Matching. These tasks provide complementary supervision for temporal perception and reasoning.

\section{Post-Training}

\subsection{Supervised Fine-Tuning and Synthetic CoT}

After multimodal pre-training, we conduct SFT to convert visual perception and cross-modal alignment into stable instruction-following behavior. This stage adapts multimodal capabilities to conversational instructions, activates perception, reasoning, long-context, and agentic abilities, and mitigates language degradation during multimodal fine-tuning.

The SFT corpus contains approximately 500B tokens and is organized around modality balance, capability coverage, and long-range modeling. It includes Text NLP, Video, Perception, Reasoning, Agent, and Long-context data. About 40\% of the corpus is text-only data, which anchors instruction following, knowledge QA, and textual reasoning.

\subsubsection{Multimodal Instruction Mixture}

To improve comprehensive instruction-following capabilities in complex multimodal scenarios, we construct a multimodal instruction mixture that covers video understanding, visual perception, cross-modal reasoning, agent-oriented tasks, and long-context modeling. The goal of this mixture is not to simply increase the proportion of visual data, but to encourage complementary capability development across perception, localization, reasoning, and task execution.

Video data are mainly used to strengthen temporal perception and evidence localization. Some samples are formulated as multiple-choice question answering tasks with clue intervals. The model is required to verify candidate temporal segments during the \texttt{\textless think\textgreater} stage, and to output both the final answer and the supporting intervals in the format \texttt{[[mm, mm], ...]}. This design encourages the model to locate key evidence from continuous video content, rather than relying only on global visual impressions.

Perception data cover OCR, document understanding, chart understanding, general visual question answering, captioning, grounding, counting, and image identification. These tasks improve the model's ability to extract, recognize, and structure fine-grained visual information. Reasoning data include K12-level exercises, STEM problems, and spatial reasoning tasks, which train the model to perform multi-step inference under both visual and textual constraints. 

Agent data consist of code reasoning and tool-use trajectories, enabling the model to learn task decomposition, execution, and feedback integration. Long-context data target long documents, long videos, multi-image inputs, and long-range question answering, with the goal of improving information retention, retrieval, and cross-segment association over extended contexts.

Overall, this instruction mixture is designed to promote capability collaboration across different data sources. Video data enhance continuous scene understanding and temporal evidence localization; perception data improve fine-grained information extraction; reasoning data strengthen cross-modal multi-step inference; agent data support task decomposition and tool interaction; and long-context data improve long-range information modeling. Meanwhile, text-only instruction data are retained to preserve general instruction following and language reasoning capabilities, reducing the risk of degrading pure language performance during multimodal training.

\subsubsection{Synthetic CoT}

Most multimodal instruction data provide only final answers, offering limited supervision for explicit reasoning and evidence aggregation. We therefore construct Synthetic CoT data from high-quality QA pairs. A strong teacher model generates reasoning traces, which are then filtered through query-level, response-level, and process-level quality checks. For mathematical tasks, a Doubt2Clean second-pass review further cleans doubtful CoT samples across 27 datasets.

The final mixture combines long \texttt{<think>} reasoning data for STEM, counting, video reasoning, and complex image-text QA with direct-answer data for general VQA, caption, OCR, and grounding. This balances explicit reasoning with concise answering and avoids unnecessary verbosity on simple perception tasks.

For continuous perception tasks, such as counting and video understanding, Synthetic CoT provides object-by-object or interval-by-interval verification chains. For example, the model may first verify local evidence in separate regions or candidate video intervals, then aggregate the evidence before producing the final answer. For STEM tasks, the reasoning trace emphasizes planning, symbolic derivation, and result verification. This gives the model learnable supervision for both ordered observation and logical reasoning. 

Beyond these offline reasoning traces, we further optimize the model's reasoning chains through strong-to-weak on-policy distillation, in which a stronger teacher supervises the student on its own on-policy rollouts.

\subsection{Reinforcement Learning}

\subsubsection{Synthetic-Data RL}
\label{sec:synthetic-data-rl}

Fine-grained multimodal reasoning requires accurate visual perception, difference localization, and structured output generation. However, natural multimodal corpora rarely provide fine-grained localization signals, structured annotations, and automatically verifiable supervision simultaneously. To address this limitation, we introduce a synthetic-data RL framework based on procedurally generated difference-recognition tasks. Given two images that differ by controlled edits, the model is required to identify the changes and report them in a predefined structured format. Since the edit operations are fully known during data generation, rewards can be computed through rule-based verification without relying on an additional learned reward model.


Each sample contains an image pair $(I_A, I_B)$, where $I_B$ is obtained from $I_A$ through a set of controlled edits $\mathcal{E}={e_1,\ldots,e_k}$. We instantiate this framework in two forms. For localization-oriented perception tasks, the model predicts normalized bounding boxes for changed regions. For structured reasoning tasks, the model outputs a domain-specific operation-set DSL corresponding to the underlying edits. In our implementation, the structured tasks cover geometry, coordinate-aware geometry, chemical formulas, and physical circuit diagrams. The synthetic setup includes both positive samples with controlled edits and negative samples with no edits, enabling the model to distinguish true changes from unchanged inputs and suppress unsupported predictions. To prevent the model from relying on trivial pixel-level differencing, we introduce difference-irrelevant re-rendering perturbations that preserve the underlying edit set, including color jitter, layout perturbation, slot shuffling, semantic no-op changes, and viewpoint variation.

For perception tasks, predicted boxes $\hat{\mathcal{B}}$ are matched to ground truth $\mathcal{B}$ using Hungarian matching based on IoU. The reward is:
\begin{equation}
R_{\mathrm{perc}} =
\lambda_F F_{\mathrm{soft}} +
\lambda_I
\frac{\sum_{m \in \mathcal{M}} u_m}
{\max(|\hat{\mathcal{B}}|,|\mathcal{B}|)}
-
\lambda_D N_{\mathrm{dup}},
\end{equation}
where $\mathcal{M}$ denotes the matched set, $u_m$ is the IoU of the $m$-th matched pair, $F_{\mathrm{soft}}$ is a soft matching-based F-score, and $N_{\mathrm{dup}}$ penalizes duplicate predictions. For negative samples, empty predictions are rewarded, while unsupported predictions are penalized. For structured domains, rewards are computed by canonicalizing the predicted and ground-truth operation sets and then performing rule-based matching. For coordinate-aware tasks, we further use soft distance-based similarity to tolerate small numerical deviations. This reward design is lightweight, scalable, and independent of learned reward models, while encouraging visual grounding, structural correctness, anti-hallucination behavior, and disciplined structured output generation.

\subsubsection{General RL}

After SFT and distillation, General RL improves reasoning ability, answer reliability, and robustness across multimodal and text-only scenarios. Unlike preference-oriented Alignment RL, General RL focuses on tasks with verifiable ground-truth answers, including general VQA, STEM, chart understanding, mathematical reasoning, logical reasoning, and text-only QA.

\paragraph{Data Construction.}
We collect multimodal and text-only QA data from open-source and in-house sources, including FineVision~(\cite{finevision}), MMK12 and MM-Eureka~(\cite{mm_eureka}), Thyme~(\cite{thyme}), mini O3~(\cite{minio3}), Open-R1~(\cite{openr1}), self-cognition data, and business-oriented data~(\cite{kuaimod}). The data pipeline includes benchmark decontamination, cross-validation with strong models and judge models, and accuracy-based filtering to remove samples already solved by the initial policy.

Benchmark decontamination removes samples that are visually or semantically too close to public evaluation instances. Cross-validation samples multiple candidate solutions and asks a judge model to select reliable positives, which are then used as reference solutions for ContextRL Reward. Accuracy-based filtering focuses RL on samples that are neither trivial nor too noisy, improving data efficiency.

\paragraph{Reward System.}
The reward system, based on Qwen2.5-VL-72B-Instruct~(\cite{qwen2_5_vl}), evaluates format validity, outcome correctness, process correctness, and consistency with verified reference solutions. Additional task-specific rewards are used for self-cognition, business tasks, and structured reasoning.

\textit{Format Reward} ensures that reasoning and final-answer fields can be parsed. \textit{Outcome Reward} checks whether the final answer matches the ground truth across heterogeneous answer formats, including options, phrases, numerical answers, and complete solutions. \textit{Process Reward} penalizes factual errors, invalid deductions, unsupported assumptions, and inconsistent intermediate steps. \textit{ContextRL Reward} ~(\cite{lu2026contextrl}) compares the sampled response with verified reference solutions to effectively reduce false positives caused by correct answers reached through flawed reasoning.

\paragraph{Training Algorithm.}
We use Group Sequence Policy Optimization (GSPO)~(\cite{zheng2025gspo}). Given query $x$, the old policy $\pi_{\theta_{\mathrm{old}}}$ samples responses $\{y_i\}_{i=1}^{G}$. The objective is:
\begin{equation}
\mathcal{J}_{\mathrm{GSPO}}(\theta) =
\mathbb{E}_{x \sim \mathcal{D}, \{y_i\}_{i=1}^{G} \sim \pi_{\theta_{\mathrm{old}}}(\cdot | x)}
\left[
\frac{1}{G}
\sum_{i=1}^{G}
\min
\left(
s_i(\theta)\hat{A}_i,
\mathrm{clip}\left(s_i(\theta), 1-\epsilon, 1+\epsilon\right)\hat{A}_i
\right)
\right],
\end{equation}
where
\begin{equation}
s_i(\theta) =
\exp
\left(
\frac{1}{|y_i|}
\sum_{t=1}^{|y_i|}
\log
\frac{
\pi_{\theta}(y_{i,t}\mid x, y_{i,<t})
}{
\pi_{\theta_{\mathrm{old}}}(y_{i,t}\mid x, y_{i,<t})
}
\right),
\end{equation}
and
\begin{equation}
\hat{A}_i =
\frac{
\mathcal{R}(x, y_i) -
\operatorname{mean}\left(\{\mathcal{R}(x, y_j)\}_{j=1}^{G}\right)
}{
\operatorname{std}\left(\{\mathcal{R}(x, y_j)\}_{j=1}^{G}\right) + \delta
}.
\end{equation}
We further over-sample responses by enlarging the generation batch size and filter groups with zero advantage variance to improve data efficiency.
This filtering keeps training focused on groups where different rollouts receive distinguishable rewards, so each update contains a stronger learning signal.

\subsubsection{Specialized RL}

Specialized RL strengthens image-text capabilities including grounding, spatial understanding, mathematical reasoning, counting, and OCR. All experts start from the same general RL checkpoint and use domain-specific data and rewards. Deterministic tasks use rule-based verifiable rewards: IoU and bipartite matching for grounding, symbolic equivalence for math, exact numeric matching for counting, and normalized text matching for OCR. Open-ended spatial reasoning uses a model judge with format constraints.

The purpose of this stage is not to train separate final models, but to obtain strong domain experts that can later contribute capability-specific supervision through distillation and capability consolidation.

\paragraph{Grounding Expert.}
For grounding tasks, the reward is designed to measure target-level localization quality while suppressing redundant predictions. We first normalize all boxes to the $[0,1000]$ coordinate range and remove duplicate predicted boxes. For multi-object grounding, predictions are matched to ground-truth boxes through Hungarian matching using IoU as the matching criterion, ensuring a one-to-one correspondence between predicted and target boxes.

Based on the matched pairs, the reward considers both the worst-case and average localization quality. The minimum IoU is used as a thresholding signal to ensure that all targets reach a basic localization quality before the sample enters the high-reward region, while the mean IoU provides continuous feedback for optimization. A duplicate-box penalty is further introduced to reduce redundant outputs and discourage box-spamming. This reward design therefore balances strict target coverage, smooth localization feedback, and output compactness.

\paragraph{Spatial Expert.}
The spatial expert targets spatial-relation understanding and embodied spatial reasoning. Unlike grounding or counting, many spatial tasks do not admit a simple deterministic criterion, since the answer may involve relative position, orientation, interaction, or scene-level commonsense. We therefore use a generative model judge to evaluate whether the response satisfies the required spatial relation and scene constraint. The judge assigns a discrete correctness score in $\{-1,0,1\}$, which is combined with a format reward to encourage both spatial correctness and legal answer structure.

\paragraph{Math Expert.}
The math expert focuses on mathematical and STEM reasoning. For problems with deterministic answers, we adopt a symbolic-equivalence reward: the model prediction and the reference answer are parsed into canonical forms, and a positive reward is assigned when they are mathematically equivalent. The reward is gated by the required \texttt{<think>} and \texttt{<answer>} structure, so that the model is encouraged to produce valid reasoning and final-answer formats. For open-ended solutions that are difficult to canonicalize, we fall back to a generative model judge. This design emphasizes answer correctness while avoiding overfitting to a specific reasoning trace.

\paragraph{Counting Expert.}
The counting expert is designed for visual counting tasks, where the target output is usually a discrete integer. Since the correctness criterion is unambiguous, we use an exact numeric-match reward under format gating. A positive reward is given only when the predicted number matches the ground-truth count. This provides a clean and deterministic learning signal, encouraging the model to align visual perception with precise numerical outputs.

\paragraph{OCR Expert.}
The OCR expert targets text recognition, document reading, and image-based text understanding. We use a normalized text-match reward, where predictions and references are compared after normalizing case, whitespace, and punctuation. This reward preserves deterministic verification while allowing minor surface-form variations. It encourages the model to improve both recognition accuracy and text completeness in OCR and document-style tasks.

\subsubsection{Video RL}

Video RL further optimizes temporal alignment, event tracking, and long-range information aggregation. Starting from the general RL checkpoint, we train on approximately 31K video samples with GSPO while freezing the visual encoder and vision-language projector.

The data cover temporal video grounding, temporal dense captioning~(\cite{vcap}), frame-level perception, video QA, temporal ordering, and event counting. TVG samples are selected from TimeIT~(\cite{timeit}), mainly from DiDeMo~(\cite{didemo}) and Charades-STA~(\cite{charades_sta}), and are rewarded by temporal IoU. Temporal dense captioning uses an LLM-as-Judge reward over subject recognition, action description, scene information, OCR text, temporal order, hallucination, and coverage. FrameForge synthetic videos provide rule-verifiable supervision for timestamp localization, counting, before/after reasoning, and co-occurrence reasoning. This stage improves general video benchmark performance by approximately 1 percentage point.

These tasks provide complementary video supervision. TVG directly optimizes temporal boundary localization, dense captioning encourages full-video coverage and faithful content organization, and FrameForge supplies low-noise frame-level signals that are difficult to obtain from natural videos.

\subsubsection{Agentic RL}

\paragraph{Shared Training Protocol.}
Agentic RL extends post-training from single-response scoring to multi-step environment interaction. It covers code, tool-use, and search tasks whose supervision comes from executable outcomes or verifiable environment states rather than static response labels. Across these domains, we use environment-grounded rewards, trajectory-level validation, and filtering to remove invalid, unfinished, low-information, or weakly verified rollouts. Completed trajectory groups are optimized with the GSPO objective introduced above, using outcome-based rewards at the trajectory level. To improve rollout utilization under long and variable interaction horizons, we use a shared colocated partial-rollout mechanism: unfinished trajectories are cached and resumed in later rollout steps, while completed groups are immediately consumed for GSPO updates. This shared protocol keeps the optimization objective and rollout scheduling consistent, while allowing each task family to define its own environment interface and reward evidence.

\paragraph{Coding RL.}
For Coding RL, we use both Online Judge and Software Engineering environments. Online Judge tasks evaluate isolated program synthesis through compilation and hidden test execution, with rewards determined by pass rate and execution failures such as time-limit or memory-limit errors. Software Engineering tasks evaluate repository-level issue resolution in containerized environments, where the model must inspect logs, run tests, edit files, and submit patches. Rewards are based on test-suite outcomes, regression checks, trajectory filtering, and auxiliary judge verification of the submitted patch. For repository-level tasks, candidate edits may further pass through a verification-and-integration protocol in which multiple reviewer agents inspect proposed changes, execution traces, and test outcomes, and a separate integration agent consolidates the reviews before final patch acceptance. This setup strengthens algorithmic correctness, repository navigation, iterative repair, and regression-safe code modification.

\paragraph{Tool Use RL.}
For tool-use RL, the model interacts with multi-turn stateful environments covering more than 150 simulated API domains. Rewards are computed from valid tool invocation, argument consistency, database or environment post-conditions, and judge-based assessment of whether the final state satisfies the user intent. Unlike code execution rewards, this objective emphasizes correct state transitions, robust recovery from tool errors, and coordination with multi-turn user instructions. Randomized tool and parameter names reduce dependence on memorized API schemas and encourage general tool-calling behavior.

\paragraph{Search RL.}
For Search RL, we train on multi-turn search tasks where the model issues retrieval actions, reads returned content, and produces a final answer after one or more retrieval rounds. These trajectories vary with the amount of query refinement, result selection, and content checking required by the task. Rewards are outcome-oriented: final-answer correctness is the primary signal, and when available we also use lightweight verification signals for intermediate search results to reduce off-track retrieval and unsupported answer synthesis. The shared partial-rollout mechanism is useful in this setting because search trajectories have variable horizons; unfinished interactions can be resumed in later rollout steps, while completed trajectory groups are used for GSPO updates.

\subsubsection{Cross-Modal Multi-Teacher On-Policy Distillation}
\label{sec:mopd}

Domain-specific post-training introduces heterogeneous capabilities from pure-text, image, video, and agent tasks. Directly mixing them may cause interference, such as overly short responses after reasoning RL or excessive tool-call formatting after agent training. To consolidate these capabilities, Keye-VL-2.0-30B-A3B uses \emph{Cross-Modal Multi-Teacher On-Policy Distillation} (MOPD).

MOPD maintains 13 RL-trained domain teachers. These teachers span the full capability spectrum, including safety, pure-text math, instruction following, code, visual STEM, OCR, grounding, counting, video, and tool use, among others. Each sample is routed to the teacher that best matches its modality and task type. Given prompt $x_i$, the student first generates an on-policy response:
\begin{align}
    y_i=(y_{i,1},\ldots,y_{i,T}) \sim \pi_\theta(\cdot \mid x_i).
\end{align}
For state $s_{i,t}=(x_i,y_{i,<t})$, the routed teacher $\pi_{\mathrm{T}}^{r(i)}$ provides token-level feedback. We use Segmented Prompt-Response Re-tokenization (SPRR) to process prompts and responses separately, ensuring strict alignment between teacher log probabilities and student response tokens.

For stable feedback, we define the top-$k$ overlap set:
\begin{align}
    \Omega_{i,t}=
    \mathrm{TopK}\left(\pi_{\mathrm{T}}^{r(i)}(\cdot \mid s_{i,t})\right)
    \cap
    \mathrm{TopK}\left(\pi_{\theta}(\cdot \mid s_{i,t})\right).
\end{align}
When $\Omega_{i,t}$ is non-empty, the raw advantage is:
\begin{align}
    A_{i,t}=
    \sum_{v\in\Omega_{i,t}}
    \bar{\pi}_{\theta}(v\mid s_{i,t})
    \left[
    \log \pi_{\mathrm{T}}^{r(i)}(v\mid s_{i,t}) -
    \log \pi_{\theta}(v\mid s_{i,t})
    \right],
\end{align}
where
\begin{align}
    \bar{\pi}_{\theta}(v\mid s_{i,t})=
    \frac{\pi_{\theta}(v\mid s_{i,t})}
    {\sum_{u\in\Omega_{i,t}}\pi_{\theta}(u\mid s_{i,t})}.
\end{align}
If $\Omega_{i,t}=\emptyset$, we set $A_{i,t}=0$.

Compared with distilling over the full vocabulary, the overlap estimator focuses supervision on the local distribution region considered plausible by both teacher and student. This avoids unstable comparisons over very low-probability tokens while keeping the training on-policy, since the feedback is computed on states actually visited by the student.

The student is optimized with an advantage-weighted policy-gradient objective:
\begin{align}
    \mathcal{L}_{\mathrm{MOPD}}=
    -\mathbb{E}
    \left[
    \frac{1}{|\mathcal{M}_i|}
    \sum_{t\in \mathcal{M}_i}
    \widehat{A}_{i,t}
    \log \pi_\theta(y_{i,t}\mid x_i,y_{i,<t})
    \right],
\end{align}
where $\mathcal{M}_i$ is the valid response-token mask. We further apply token-category-aware advantage scaling to down-weight formatting tokens and up-weight perception or reasoning tokens. For long-form generation, repetition collapse is localized at position $\tau_i$ and penalized only after the collapse point:
\begin{align}
    \widehat{A}_{i,t}=
    \widetilde{A}_{i,t} -
    \lambda_{\mathrm{rep}}\cdot \mathbb{I}[t\ge \tau_i].
\end{align}

Together, dynamic teacher routing, SPRR alignment, top-$k$ overlap estimation, token-category scaling, and localized repetition penalties allow MOPD to consolidate heterogeneous RL teachers without forcing all domains into a single response style.

\section{Efficient Training and Inference Infrastructure}

\subsection{Pre-Training Systems}

\subsubsection{ViT--LM Heterogeneous Parallelism and Load Balancing}

To remove I/O bottlenecks from video decoding and frame sampling, we introduce \textbf{ExtraIO}, a horizontally scalable I/O service decoupled from training through an asynchronous pipeline. We further co-design ViT--LM heterogeneous parallelism and two-level load balancing for long-video and variable-length workloads.

\paragraph{ViT--LM Heterogeneous Parallelism.}
The ViT and LM are co-located on the same GPU group, but each module adopts its own parallel sharding strategy. This avoids the imbalance caused by binding the ViT to LM PP0. A recompute-or-offload strategy reduces ViT activation memory to nearly zero, leaving more memory for long sequences.

\paragraph{Load Balancing.}
Because the ViT processes image/frame-level visual tokens while the LM processes sample-level multimodal sequences, both visual-token ratios and sequence lengths fluctuate. We balance load at the multimodal-token level and the LM-sample level, equalizing compute and memory across ViT DP and LM DP/PP. This improves end-to-end training throughput by approximately \textbf{20\%}.

This design is particularly useful for video-heavy batches, where a few long samples can otherwise create severe bubbles and leave a large fraction of devices underutilized.

\subsubsection{DSA Optimization for Variable-Length Sequences}

We optimize DSA with \textbf{FlashInfer}~(\cite{flashinfer}) and \textbf{TileLang}~(\cite{tilelang}), achieving more than a \textbf{2$\times$} speedup over a baseline adapted from open-source code.

\paragraph{Top-$k$ Memory Optimization.}
Under packing, the original indexer score has shape $T \times T$, causing memory waste because samples are mutually invisible. We reduce score storage to $T \times \text{max\_seq}$ and use \texttt{flashinfer.top\_k\_ragged\_transform} to compute only over valid KV regions. A chunked indexer is used as a memory-bound fallback for extreme cases.

\paragraph{Short-Sequence Optimization.}
During long-sequence SFT, many samples are short. When the positional index satisfies $i < \text{top}k$, the indexer backward and sparse-attention kernels iterate only over causally attendable KV entries instead of scanning a fixed top-$k$ range, bringing a \textbf{1.5$\times$} end-to-end speedup.

\paragraph{Indexer Loss.}
Instead of storing full indexer and attention scores across layers, we keep only post-top-$k$ indexer scores during forward and recover attention scores inside the sparse-attention backward kernel. This reuses FlashAttention-style backward recomputation~(\cite{flashattention}), avoids additional compute, and releases per-layer intermediates early.

The key observation is that sparse-attention backward already recomputes the post-softmax attention matrix. By extracting and reducing the needed scores inside the backward kernel, the indexer loss can be computed without materializing layer-wise $T \times T$ intermediates.

\subsection{Post-Training Systems}

\subsubsection{DSA Adaptation for RL}

For RL training with DSA, we focus on consistency and memory. To avoid mismatch between training and inference Top-$k$ results, we use deterministic Top-$k$ computation. \texttt{flashinfer.topk} replaces \texttt{torch.topk}, achieving a \textbf{2--3$\times$} speedup while preserving determinism. For variable-length RL batches, a chunked DSA indexer partitions the score matrix along the sequence dimension, performs Top-$k$ chunk by chunk, and aggregates the results to reduce peak memory.

\subsubsection{On-Policy Distillation System}

The OPD system supports heterogeneous multi-expert teacher scheduling, multimodal alignment, and Top-$k$ distillation. Samples are routed to domain-specific teachers such as Math, Grounding, OCR, and LAN-Instruct, while domains sharing weights reuse the same server instance. To align student and teacher preprocessing, the teacher side reconstructs sequences with its native processor and verifies visual-token counts, positional encodings, and multimodal-text boundaries. Top-$k$ distillation supports Overlap, Student-Only, and Teacher-on-Student modes.

Strict multimodal alignment is essential because even a small mismatch in image token count, video frame sampling, chat template, or mRoPE position can move the KL signal to the wrong response token. The system therefore treats alignment verification as part of the training pipeline rather than an offline debugging step.

\subsection{Efficient Inference for GQA+DSA}

For ultra-long video inference, we introduce three optimizations.

\begin{itemize}
    \item \textbf{Chunk ViT:} video frames are split into chunks, processed sequentially by the ViT, and then merged, reducing peak memory without changing model outputs.
    \item \textbf{Sparse Attention Optimization:} adjacent queries often select similar Top-$k$ KV sets. We deduplicate Top-$k$ sets across adjacent queries and use an MMA Thread Layout-Aware Mask inside the attention kernel. With a 128K context and $\text{top}k=2048$, 16 adjacent queries require only about 8K effective KV tokens.
    \item \textbf{Decode Optimization:} DSA-specific decode optimizations reduce prefill cost by over \textbf{3$\times$} and decode cost by over \textbf{5$\times$} compared with full attention under a 128K context.
\end{itemize}

\begin{figure*}[!t]
\centering
\includegraphics[width=1.0\textwidth]{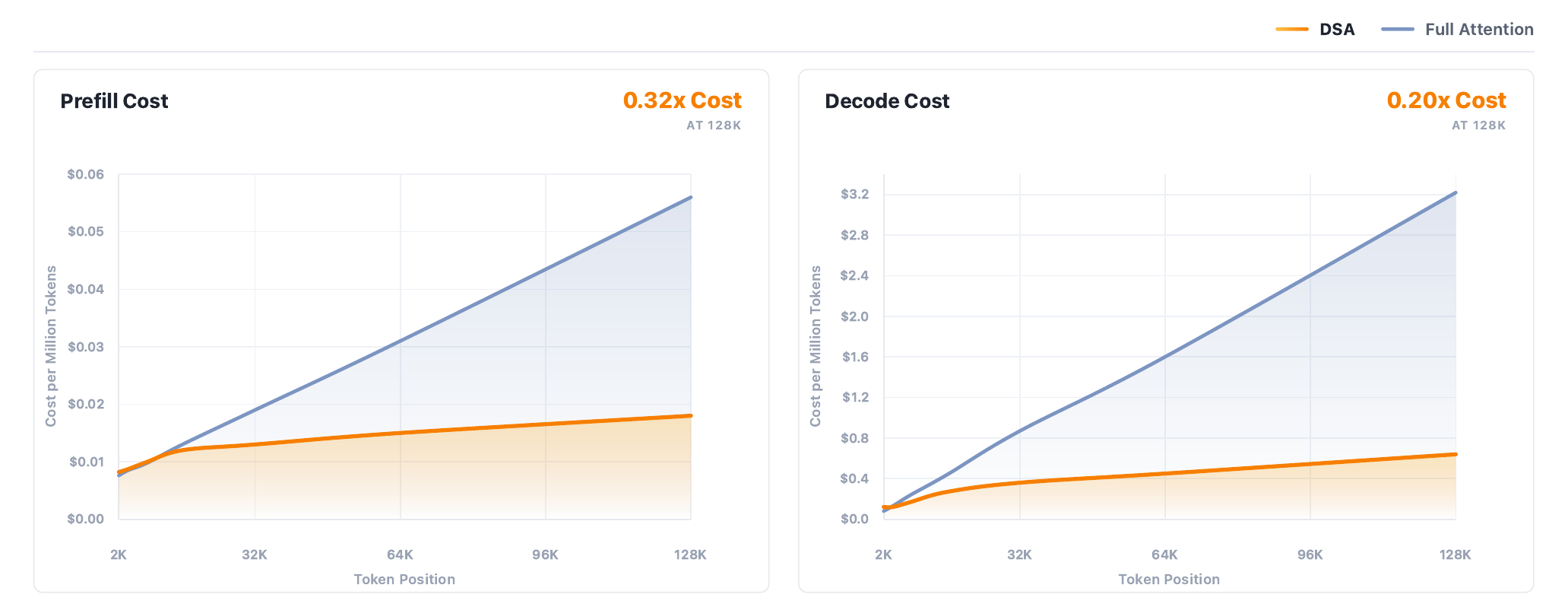}
\caption{\textbf{Inference cost of Keye-VL-2.0-30B-A3B.} DSA-specific prefill and decode optimizations reduce the cost of ultra-long video inference relative to dense attention under the same H800 pricing assumption.}
\label{fig:inference_cost}
\end{figure*}

Experiments on H800 GPUs, assuming \$2 per GPU-hour, demonstrate the inference efficiency shown in Figure~\ref{fig:inference_cost}.

\FloatBarrier

\section{Comprehensive Evaluation}

\begin{figure*}[!t]
\centering
\includegraphics[width=0.96\textwidth]{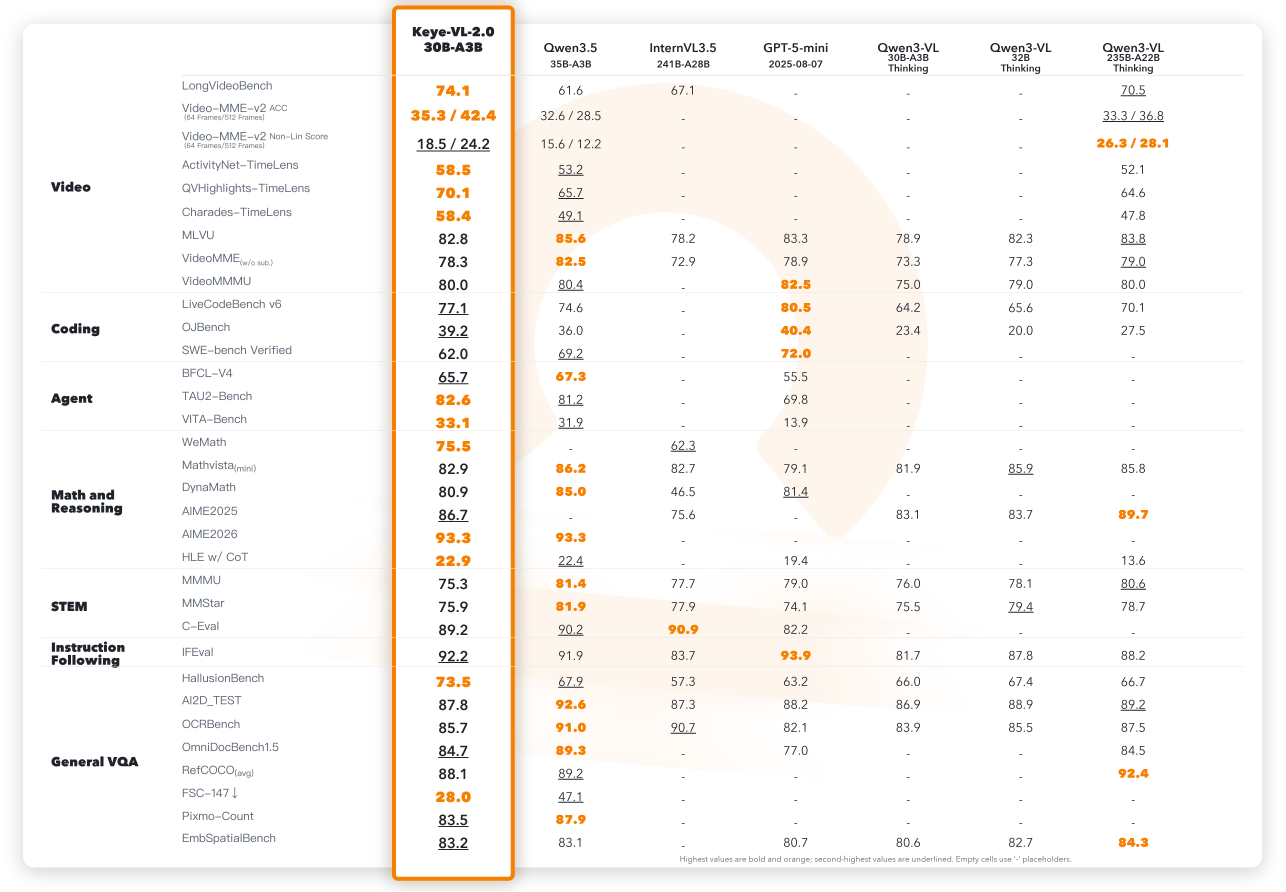}
\caption{\textbf{Overall evaluation summary of Keye-VL-2.0-30B-A3B.} The figure summarizes representative results across video understanding, coding, agentic tool use, mathematical and scientific reasoning, instruction following, and general vision-language benchmarks. Orange scores mark leading results in each row, and ``--'' indicates unavailable or not directly comparable scores. Higher is better unless otherwise specified by the corresponding benchmark; detailed benchmark descriptions and citations are provided in the subsections below.}
\label{fig:overall_eval}
\end{figure*}

Before presenting benchmark-specific analyses, Figure~\ref{fig:overall_eval} provides a compact overview of Keye-VL-2.0-30B-A3B across video understanding, coding, agentic tool use, mathematical and scientific reasoning, instruction following, and general vision-language benchmarks. We compare with representative open-source and closed-source baselines of similar and larger scales, including Qwen3.5~(\cite{qwen3.5}), InternVL3.5~(\cite{internvl35}), GPT-5-mini~(\cite{openai2025gpt5developers}), and Qwen3-VL~(\cite{qwen3vl}). The following subsections give the benchmark definitions, evaluation protocols, and detailed results behind this overall view.

\subsection{Video Understanding}
\label{sec:video_eval}

We evaluate Keye-VL-2.0-30B-A3B on three complementary aspects of video understanding: long-video comprehensive understanding, fine-grained temporal grounding, and video knowledge acquisition.

\subsubsection{Benchmarks}

\paragraph{Long-Video Comprehensive Understanding.}
Video-MME-v2~(\cite{fu2026videomme2}) evaluates omni-modal information aggregation, long-range temporal understanding, and complex reasoning. It reports both average accuracy and a group-based non-linear score $(N/4)^2$, where $N$ is the number of correct answers in a group of four correlated questions. LongVideoBench~(\cite{longvideobench}) evaluates long-context video-language reasoning with referring queries. MLVU~(\cite{mlvu}) covers multi-task long-video understanding from 3-minute to 2-hour videos. Video-MME~(\cite{video_mme}) evaluates short, medium, and long videos; we report the no-subtitle setting.

These benchmarks together test whether a model can retrieve relevant segments from long videos, maintain event order, integrate cross-segment evidence, and answer questions without relying on subtitle shortcuts.

\paragraph{Temporal Grounding.}
We use the TimeLens framework~(\cite{zhang2025timelens}), which re-annotates ActivityNet Captions, QVHighlights, and Charades-STA to reduce annotation noise. We report mIoU on ActivityNet-TimeLens, QVHighlights-TimeLens, and Charades-TimeLens.

The cleaned annotations make TimeLens a more reliable probe of fine-grained temporal alignment than the original legacy annotations, where noisy boundaries can distort model rankings.

\paragraph{Video Knowledge Acquisition.}
Video-MMMU~(\cite{videommmu}) evaluates whether a model can learn and apply domain knowledge from instructional videos across 30 sub-disciplines and 6 major fields.

Its questions cover perception, comprehension, and adaptation levels, requiring the model to identify information in the lecture, understand the underlying concept, and apply it to new scenarios.

\subsubsection{Results}

\begin{table}[t]
\centering
\caption{Video understanding evaluation of Keye-VL-2.0-30B-A3B against representative baselines. \textbf{Bold} marks the best result per row, \underline{underline} marks the second-best, and ``--'' indicates an unavailable comparable score.}
\label{tab:video_eval}
\setlength{\tabcolsep}{3pt}
\renewcommand{\arraystretch}{1.15}
\resizebox{\textwidth}{!}{%
\begin{tabular}{llccccccc}
\toprule
\textbf{Category} & \textbf{Benchmark} &
\makecell{\textbf{Keye-VL-2.0}\\\textbf{30B-A3B}} &
\makecell{Qwen3.5\\35B-A3B} &
\makecell{InternVL3.5\\241B-A28B} &
\makecell{GPT-5-mini\\2025-08-07} &
\makecell{Qwen3-VL\\30B-A3B\\Thinking} &
\makecell{Qwen3-VL\\32B\\Thinking} &
\makecell{Qwen3-VL\\235B-A22B\\Thinking} \\
\midrule
\multirow{5}{*}{\makecell[l]{Long-Video\\Comprehensive}}
 & LongVideoBench                         & \textbf{74.1}        & 61.6        & 67.1 & --   & --   & --   & \underline{70.5} \\
 & Video-MME-v2 ACC (64 / 512 frames)     & \textbf{35.3 / 42.4} & 32.6 / 28.5 & --   & --   & --   & --   & \underline{33.3 / 36.8} \\
 & Video-MME-v2 Non-Lin (64 / 512 frames) & \underline{18.5 / 24.2} & 15.6 / 12.2 & -- & -- & -- & -- & \textbf{26.3 / 28.1} \\
 & MLVU                                   & 82.8                 & \textbf{85.6} & 78.2 & 83.3 & 78.9 & 82.3 & \underline{83.8} \\
 & Video-MME (w/o sub.)                   & 78.3                 & \textbf{82.5}    & 72.9 & 78.9 & 73.3 & 77.3 & \underline{79.0} \\
\midrule
\multirow{3}{*}{\makecell[l]{Temporal Grounding\\(TimeLens)}}
 & ActivityNet-TimeLens                   & \textbf{58.5} & 53.2             & -- & -- & -- & -- & 52.1 \\
 & QVHighlights-TimeLens                  & \textbf{70.1} & \underline{65.7} & -- & -- & -- & -- & 64.6 \\
 & Charades-TimeLens                      & \textbf{58.4} & 49.1             & -- & -- & -- & -- & 47.8 \\
\midrule
\makecell[l]{Video Knowledge\\Acquisition}
 & Video-MMMU                             & 80.0 & \underline{80.4} & -- & \textbf{82.5} & 75.0 & 79.0 & 80.0 \\
\bottomrule
\end{tabular}%
}
\end{table}

Keye-VL-2.0 achieves the best result on LongVideoBench and Video-MME-v2 accuracy, and remains competitive on mature benchmarks such as MLVU and Video-MME. On Video-MME-v2, its strong accuracy under both 64-frame and 512-frame settings indicates that the model benefits from denser visual context without losing long-range aggregation ability. It also achieves the best mIoU on all three TimeLens subsets, validating the effectiveness of scene-wise dense captions, diverse TVG data, and tIoU-centered Video RL. In addition, for the TimeLens benchmark, the score of Qwen3-VL-235B-A22B Thinking is directly taken from the paper, where the evaluation was conducted with 2 FPS frame sampling. For Qwen3.5-35B-A3B, there is currently no official TimeLens score available. Considering that previous Qwen-series evaluations on the TVG benchmark adopted a 4 FPS setting, we evaluate the main comparison models, including Qwen3.5 35B-A3B, Gemini, and Keye-VL-2.0, using dense frame sampling at 4 FPS for a fair comparison. On Video-MMMU, it reaches $80.0$, matching strong open-source baselines and approaching the closed-source GPT-5-mini.

\subsection{Agentic Capability Evaluation}

\subsubsection{Code Agent Evaluation}

We evaluate coding and software-engineering capabilities on LiveCodeBench v6~(\cite{livecodebench}), OJBench~(\cite{ojbench}), and SWE-bench Verified~(\cite{swe_bench_verified}). LiveCodeBench v6 provides contamination-resistant programming evaluation, OJBench tests online-judge-style algorithmic correctness, and SWE-bench Verified evaluates repository-level issue resolution.

\begin{table}[t]
\centering
\caption{Code agent evaluation of Keye-VL-2.0-30B-A3B against representative baselines. \textbf{Bold} marks the best result per row, \underline{underline} marks the second-best, and ``--'' indicates an unavailable comparable score.}
\label{tab:code_agent_eval}
\setlength{\tabcolsep}{3pt}
\renewcommand{\arraystretch}{1.15}
\resizebox{\textwidth}{!}{%
\begin{tabular}{lccccccc}
\toprule
\textbf{Benchmark} &
\makecell{\textbf{Keye-VL-2.0}\\\textbf{30B-A3B}} &
\makecell{Qwen3.5\\35B-A3B} &
\makecell{InternVL3.5\\241B-A28B} &
\makecell{GPT-5-mini\\2025-08-07} &
\makecell{Qwen3-VL\\30B-A3B\\Thinking} &
\makecell{Qwen3-VL\\32B\\Thinking} &
\makecell{Qwen3-VL\\235B-A22B\\Thinking} \\
\midrule
LiveCodeBench v6     & \textbf{64.2}    & \underline{62.8} & -- & 51.5 & -- & -- & -- \\
OJBench              & \textbf{71.5}    & \underline{70.2} & -- & 58.7 & -- & -- & -- \\
SWE-bench Verified   & \underline{62.0} & \textbf{63.5}    & -- & 55.5 & -- & -- & -- \\
\bottomrule
\end{tabular}%
}
\end{table}

Keye-VL-2.0-30B-A3B achieves $64.2$ on LiveCodeBench v6 and $71.5$ on OJBench, showing strong algorithmic reasoning and execution-based self-correction. It also obtains a competitive $62.0$ on SWE-bench Verified, suggesting that the model transfers part of its coding ability from isolated algorithmic problems to repository-level software engineering.

\subsubsection{Tool-Use Evaluation}

We evaluate function calling and multi-turn tool use on BFCL-V4~(\cite{bfcl}), $\tau^2$-Bench~(\cite{tau2bench}), and VitaBench~(\cite{vitabench}). These benchmarks cover single-turn and multi-turn function calling, dual-control stateful interaction, and complex life-service scenarios with heterogeneous tools.

\begin{table}[t]
\centering
\caption{Tool-use and function-calling evaluation of Keye-VL-2.0-30B-A3B against representative baselines. \textbf{Bold} marks the best result per row, \underline{underline} marks the second-best, and ``--'' indicates an unavailable comparable score.}
\label{tab:tool_use_eval}
\setlength{\tabcolsep}{3pt}
\renewcommand{\arraystretch}{1.15}
\resizebox{\textwidth}{!}{%
\begin{tabular}{lccccccc}
\toprule
\textbf{Benchmark} &
\makecell{\textbf{Keye-VL-2.0}\\\textbf{30B-A3B}} &
\makecell{Qwen3.5\\35B-A3B} &
\makecell{InternVL3.5\\241B-A28B} &
\makecell{GPT-5-mini\\2025-08-07} &
\makecell{Qwen3-VL\\30B-A3B\\Thinking} &
\makecell{Qwen3-VL\\32B\\Thinking} &
\makecell{Qwen3-VL\\235B-A22B\\Thinking} \\
\midrule
BFCL-V4        & \underline{65.7} & \textbf{67.3}    & -- & 55.5 & -- & -- & -- \\
$\tau^2$-Bench & \textbf{82.6}    & \underline{81.2} & -- & 69.8 & -- & -- & -- \\
VitaBench      & \textbf{33.1}    & \underline{31.9} & -- & 13.9 & -- & -- & -- \\
\bottomrule
\end{tabular}%
}
\end{table}

Keye-VL-2.0-30B-A3B achieves the best results on $\tau^2$-Bench and VitaBench and ranks second on BFCL-V4. These results indicate strong tool selection, parameter filling, state tracking, and recovery behavior in multi-turn environments.

\subsection{General Vision-Language Evaluation}

We evaluate general vision-language capabilities on perception- and reasoning-oriented benchmarks.

Perception-oriented benchmarks cover OCR and document understanding, grounding, counting, spatial understanding, and hallucination resistance. OCRBench~(\cite{ocrbench}) and OmniDocBench~(\cite{omnidocbench}) focus on text-centric visual understanding and document intelligence. RefCOCO~(\cite{refcoco}) evaluates referring-expression grounding, FSC-147~(\cite{fsc_147}) and PixMoCount~(\cite{pixmo}) measure counting, EmbSpatial-Bench~(\cite{embspatialbench}) evaluates egocentric spatial relations, and HallusionBench~(\cite{hallusionbench}) diagnoses perceptual reliability.

Reasoning-oriented benchmarks cover visual mathematics, dynamic reasoning robustness, expert-level multimodal understanding, and vision-indispensable reasoning. WeMath~(\cite{wemath}), MathVista~(\cite{mathvista}), and DynaMath~(\cite{dynamath}) evaluate complementary aspects of visual mathematical reasoning, while MMMU~(\cite{mmmu}) and MMStar~(\cite{mm_star}) test expert-level multimodal reasoning under reduced text-only shortcut opportunities. Overall, Keye-VL-2.0-30B-A3B maintains strong general multimodal performance while showing particularly strong results in long-video understanding, temporal grounding, hallucination resistance, and visual mathematical reasoning.

\section{Conclusion and Future Work}

We presented Kwai Keye-VL-2.0-30B-A3B, an open-source 30B-class MoE multimodal foundation model with only 3B active parameters. By bringing DeepSeek Sparse Attention into a GQA-based multimodal backbone, Keye-VL-2.0 extends effective context modeling to 256K tokens and makes hour-level video understanding practical under controllable training and inference cost. Together with native-resolution visual encoding, unified image-video processing, ViT--LM heterogeneous parallelism, DSA kernels, and Chunk ViT inference, the system is designed not only to improve benchmark accuracy, but also to support deployable long-video applications.

The post-training pipeline further addresses the capability-conflict problem that emerges when perception, reasoning, long-context understanding, and agentic behaviors are optimized together. Cross-Modal Multi-Teacher On-Policy Distillation, Context-RL, Video-RL, and specialized domain RL allow heterogeneous teachers and reward signals to be consolidated into a single MoE model without sacrificing core reasoning ability. Evaluations show that Keye-VL-2.0 achieves leading performance at its scale on long-video comprehension and fine-grained temporal localization, while remaining competitive on code, tool-use, OCR, document understanding, visual mathematics, and hallucination-resistance benchmarks. Overall, the results indicate that sparse long-context modeling and carefully staged multimodal RL can be combined in a single deployable MoE system without trading away general reasoning ability.

Future work will move beyond leaderboard-oriented optimization toward deeper deployment in real business scenarios. First, we will further integrate fine-grained long-video perception and image-text understanding into core product pipelines, including generative recommendation, content ecosystem governance, and commercial targeting. In these settings, Keye-VL is expected to provide denser semantic signals for recommendation matching, content quality assessment, and fine-grained advertising labels, turning multimodal understanding into measurable product and business value. Second, we will develop Video~$\times$~Agent workflows that combine precise multimodal understanding with automated orchestration. The model will evolve from passively understanding video content to actively coordinating production loops, including large-scale video retrieval, highlight segment extraction, automated editing and packaging, and marketing-copy generation for creator and commercial scenarios. Third, using Keye-VL-2.0-30B-A3B as a validated foundation, we will continue strengthening the underlying infrastructure from DSA-based compute optimization, scalable data flywheels, and Context-RL post-training toward native multimodal modeling and deeper end-to-end fusion. In this direction, benchmark gains serve as diagnostics rather than the final objective; the long-term goal is to turn long-context multimodal intelligence into reliable, scalable infrastructure for real-world applications.

\clearpage
\bibliographystyle{colm2024_conference}
\bibliography{colm2024_conference}

\clearpage
\appendix

\newpage
\section{Case Study}

The following qualitative examples show the final user-facing responses of Keye-VL-2.0 on representative text, image, video, and agentic service tasks. Since Keye-VL-2.0 uses a thinking-oriented policy by default, we omit internal reasoning traces and do not display separate thinking or answer tags in the case study.

\subsubsubsection{Case I: Logical Constraint Solving}
\label{text_cases}

\begin{caseprompt}
\textbf{The Protocol of the Corrupted AI.}
A cybersecurity analyst must override a rogue AI system by entering five unique Access Keys into five slots. Each Access Key contains exactly one letter from \(\{A,B,C,D,E\}\) and one digit from \(\{1,2,3,4,5\}\). No letter or digit may repeat. Five intercepted nodes describe the sequence, but exactly one node is corrupted: every statement in the corrupted node is false, while every statement in each intact node is true.

\begin{description}[leftmargin=1.8em,itemsep=0.2em,topsep=0.3em]
  \item[Node 1.] \(A\) is immediately to the left of the slot containing digit \(5\). Digit \(3\) is located in Slot 5.
  \item[Node 2.] \(C\) is paired with an even digit. The slot containing \(E\) is exactly two slots to the right of the slot containing digit \(4\).
  \item[Node 3.] Digit \(2\) is adjacent to the slot containing \(D\). \(B\) is neither in Slot 1 nor Slot 5.
  \item[Node 4.] The sum of the digits in Slot 1 and Slot 2 equals the digit in Slot 3. The digit paired with \(A\) is strictly less than the digit paired with \(B\).
  \item[Node 5.] \(D\) is somewhere to the left of \(C\). Digit \(1\) is paired directly with \(E\). Slot 4 contains an odd digit.
\end{description}

\textbf{Mission.} Identify the corrupted node and determine the exact letter-digit sequence for all five slots.
\end{caseprompt}

\begin{caseresponse}
\textbf{Corrupted node.} Node 5 is the corrupted node. All statements in Node 5 are false, while all statements in Nodes 1--4 are true.

\textbf{Exact sequence.}
\begin{center}
\begin{tabular}{ccc}
\toprule
\textbf{Slot} & \textbf{Letter} & \textbf{Digit} \\
\midrule
1 & C & 4 \\
2 & A & 1 \\
3 & E & 5 \\
4 & B & 2 \\
5 & D & 3 \\
\bottomrule
\end{tabular}
\end{center}

\textbf{Verification.}
\begin{center}
\begin{tabular}{p{0.13\linewidth}p{0.77\linewidth}}
\toprule
\textbf{Node} & \textbf{Status and evidence} \\
\midrule
Node 1 & True. \(A\) in Slot 2 is immediately to the left of digit \(5\) in Slot 3; digit \(3\) is in Slot 5. \\
Node 2 & True. \(C\) is paired with digit \(4\), which is even; \(E\) in Slot 3 is exactly two slots to the right of digit \(4\) in Slot 1. \\
Node 3 & True. Digit \(2\) in Slot 4 is adjacent to \(D\) in Slot 5; \(B\) is in Slot 4, so it is neither in Slot 1 nor Slot 5. \\
Node 4 & True. The Slot 1 and Slot 2 digits sum to the Slot 3 digit: \(4+1=5\). The digit paired with \(A\) is \(1\), which is less than the digit paired with \(B\), \(2\). \\
Node 5 & False. \(D\) is in Slot 5 and \(C\) is in Slot 1, so \(D\) is not left of \(C\). Digit \(1\) is paired with \(A\), not \(E\). Slot 4 contains digit \(2\), which is even rather than odd. \\
\bottomrule
\end{tabular}
\end{center}

The assignment satisfies the uniqueness constraint for all letters and digits and leaves exactly one corrupted node.
\end{caseresponse}

\captionof{figure}{\textbf{Text case for logical constraint solving.} Keye-VL-2.0 solves a multi-constraint symbolic puzzle by identifying the single corrupted evidence node, constructing a valid slot assignment, and verifying every statement against the final sequence.}
\label{fig:corrupted_ai_protocol_case}

\clearpage

\subsubsubsection{Case II: Spatial Reasoning in an Indoor Layout}
\label{image_cases}

\begin{center}
\CaseInputTitle{Input Image}\\[0.4em]
\IfFileExists{arXiv/cases/figs/room_layout_spatial_reasoning.png}{%
  \includegraphics[width=0.72\linewidth]{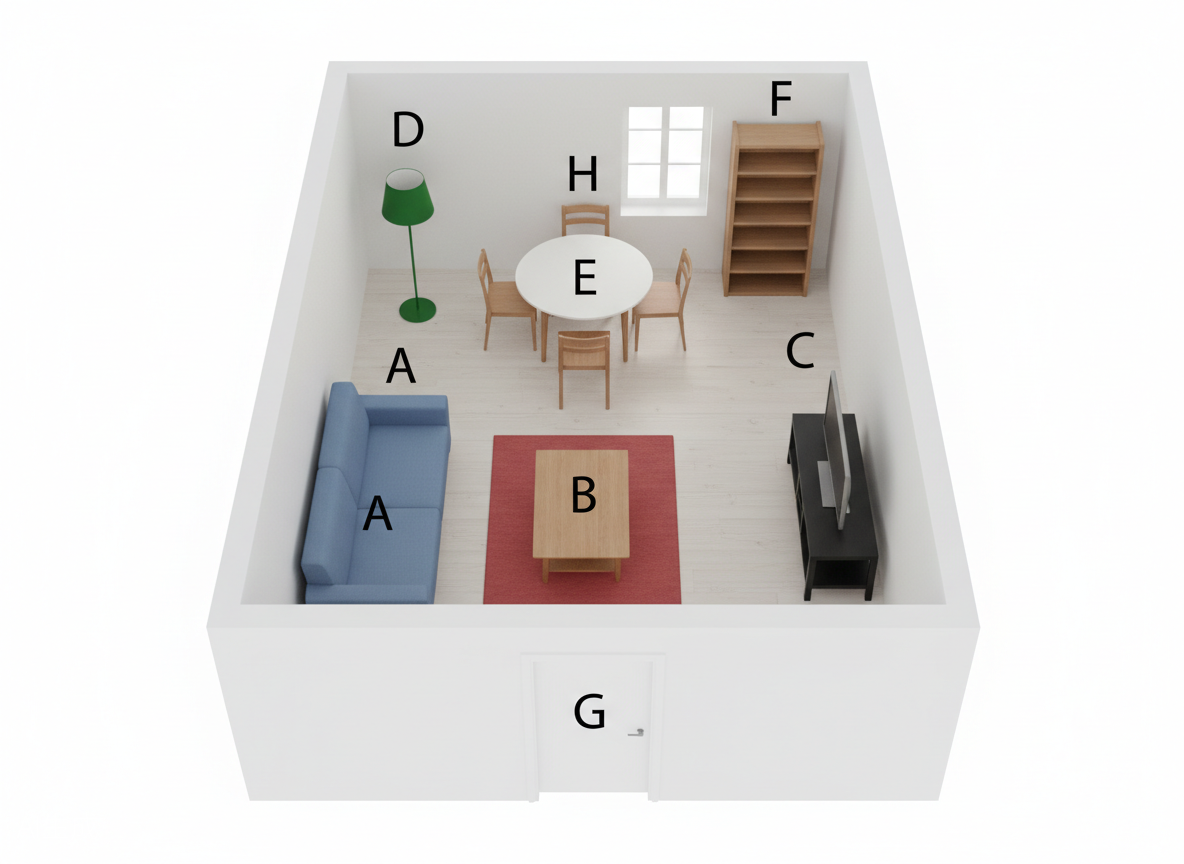}%
}{%
  \fbox{\begin{minipage}[c][5.2cm][c]{0.72\linewidth}
    \centering
    Room layout image placeholder\\
    arXiv/cases/figs/room\_layout\_spatial\_reasoning.png
  \end{minipage}}%
}
\end{center}

\begin{caseprompt}
The image is a top-down view of a room, with each piece of furniture labeled by a letter. Please answer the following questions:
\begin{enumerate}[leftmargin=1.6em,itemsep=0.15em,topsep=0.25em]
  \item Which direction is \textbf{A (the sofa)} facing?
  \item Which direction is \textbf{C (the TV)} facing?
  \item Are \textbf{A (the sofa)} and \textbf{C (the TV)} facing each other?
  \item When entering the room through \textbf{G (the door)}, what furniture is on your \textbf{left-hand side} and what furniture is on your \textbf{right-hand side}?
  \item What is the relative spatial relationship between \textbf{D (the floor lamp)} and \textbf{H (the window)}?
  \item If \textbf{B (the coffee table)} is moved next to \textbf{E (the dining table)}, in which direction should it be moved?
  \item In which corner of the room is \textbf{F (the bookshelf)} located? Describe its position relative to the other furniture.
\end{enumerate}
Please answer based on the spatial layout shown in the image, paying close attention to orientations, directions, and relative positions.
\end{caseprompt}

\begin{caseresponse}
\textbf{Directions of A and C.} A (sofa) faces toward the center of the room, i.e., to the right. C (TV) faces toward the center of the room, i.e., to the left. Therefore, A and C are facing each other, forming a typical living-room seating arrangement.

\textbf{Left and right when entering from G.} When standing at G and entering the room, the furniture on the left-hand side is the sofa A, while the furniture on the right-hand side is the TV C.

\textbf{Relative position of D and H.} D (floor lamp) is to the left of H (window), namely on the west side of the window. The floor lamp is positioned left and slightly forward of the window, and both are near the north wall.

\textbf{Moving B next to E.} B (coffee table) is currently south of E (dining table), or below it in the top-down layout. To move B next to the dining table, it should be moved north.

\textbf{Location of F.} F (bookshelf) is located in the northeast corner of the room. Relative to the other furniture, it is east of the window H, north of the TV C, and northeast of the dining table E. It is positioned near the intersection of the north and east walls.

Overall, this layout shows a standard living-room configuration: the sofa and TV face each other to form a seating area, the dining table is in the middle-north section, the window and bookshelf are on the north wall, and the door is on the south wall.
\end{caseresponse}

\captionof{figure}{\textbf{Image case for spatial layout understanding.} Given a labeled top-down indoor scene, Keye-VL-2.0 identifies object orientations, egocentric left-right relations, furniture positions, and the direction needed to move an object.}
\label{fig:room_spatial_reasoning_case}

\clearpage

\subsubsubsection{Case III: Anatomical Reasoning and Error Detection}

\begin{center}
\CaseInputTitle{Input Image}\\[0.4em]
\IfFileExists{arXiv/cases/figs/heart_circulation_diagram.png}{%
  \includegraphics[width=0.76\linewidth]{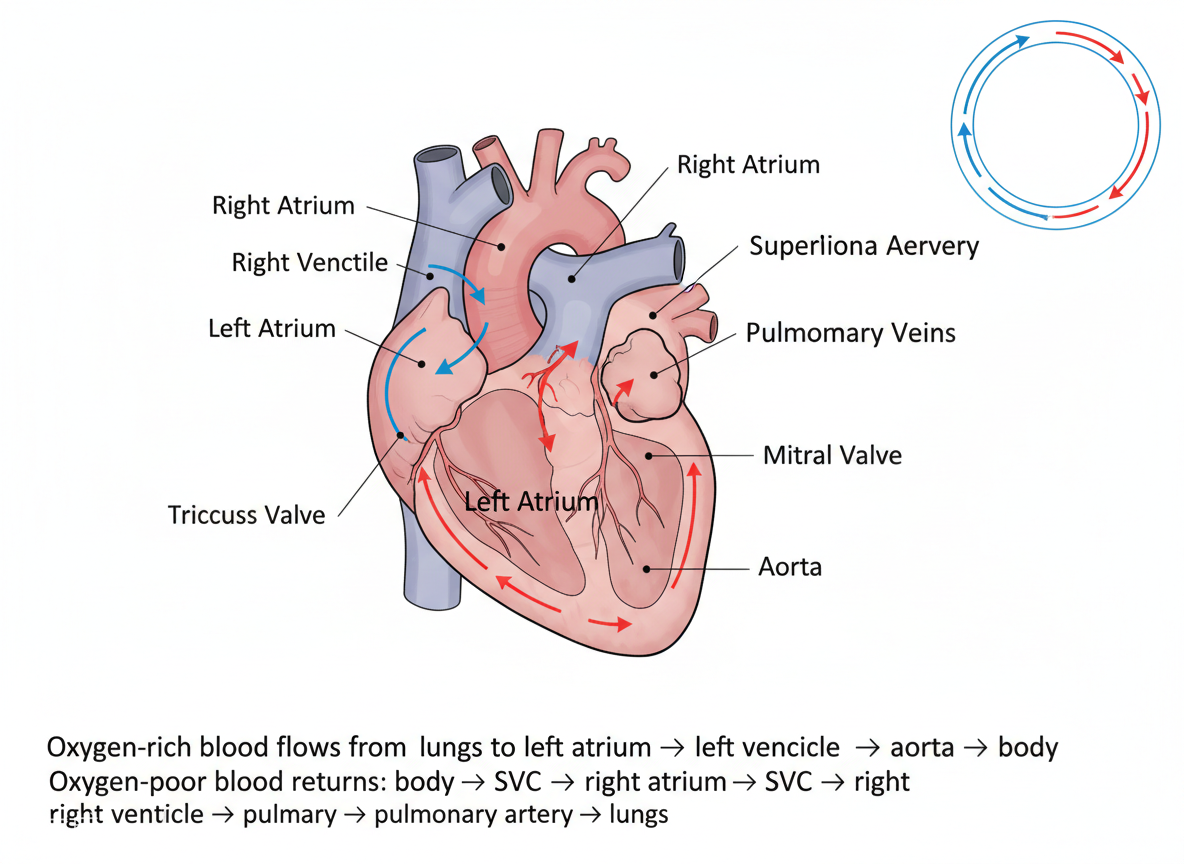}%
}{%
  \fbox{\begin{minipage}[c][5.2cm][c]{0.76\linewidth}
    \centering
    Heart circulation diagram placeholder\\
    arXiv/cases/figs/heart\_circulation\_diagram.png
  \end{minipage}}%
}
\end{center}

\begin{caseprompt}
This image is an anatomical diagram of the human heart and a schematic of blood circulation. Please answer the following questions:
\begin{itemize}[leftmargin=1.6em,itemsep=0.15em,topsep=0.25em]
  \item What are the four chambers of the heart? Please list them according to the labels in the diagram.
  \item What do the blue and red arrows in the diagram represent, respectively? Hint: it is related to blood oxygen content.
  \item After blood flows from the lungs into the heart, which chambers and blood vessels does it sequentially pass through before being delivered to the rest of the body?
  \item Where is the ``Mitral Valve'' labeled in the diagram located? Which two chambers does it connect?
  \item Based on the text description in the diagram, point out one scientifically incorrect statement and correct it.
\end{itemize}
Please base your answers on the labels and text information provided in the image.
\end{caseprompt}

\begin{caseresponse}
\textbf{Four chambers.} According to the labels in the diagram, the four chambers of the heart are Right Atrium, Right Ventricle, Left Atrium, and Left Ventricle.

\textbf{Meaning of blue and red arrows.} Blue arrows represent the flow of oxygen-poor blood with low oxygen content. Red arrows represent the flow of oxygen-rich blood with high oxygen content. This is consistent with the diagram text describing oxygen-poor blood returning and oxygen-rich blood flowing from the lungs.

\textbf{Pathway from the lungs to the body.} The sequence is: lungs \(\rightarrow\) left atrium \(\rightarrow\) left ventricle \(\rightarrow\) aorta \(\rightarrow\) entire body.

\textbf{Position of the Mitral Valve.} The Mitral Valve is located between the left atrium and the left ventricle. It connects the Left Atrium and the Left Ventricle.

\textbf{Scientific error and correction.} The path description for oxygen-poor blood states: body \(\rightarrow\) SVC \(\rightarrow\) right atrium \(\rightarrow\) SVC \(\rightarrow\) right ventricle. This is incorrect because the SVC is used twice, and it carries blood from the body to the right atrium rather than from the right atrium to the right ventricle. The corrected pathway should be: body \(\rightarrow\) SVC \(\rightarrow\) right atrium \(\rightarrow\) right ventricle \(\rightarrow\) pulmonary artery \(\rightarrow\) lungs.

Additionally, there is a spelling error in the diagram: ``pulmary'' should be ``pulmonary.''
\end{caseresponse}

\captionof{figure}{\textbf{Image case for anatomical diagram understanding.} Given a labeled heart diagram with blood-flow arrows and explanatory text, Keye-VL-2.0 identifies cardiac chambers, interprets oxygen-rich and oxygen-poor circulation, localizes the mitral valve, reconstructs the pulmonary-to-systemic pathway, and detects an incorrect textual description.}
\label{fig:heart_circulation_case}

\clearpage

\subsubsubsection{Case IV: Scene-by-Scene Historical Video Understanding}
\label{case_video_history}

\begin{center}
\CaseInputTitle{Input Video}\\[0.4em]
\includegraphics[width=0.88\linewidth]{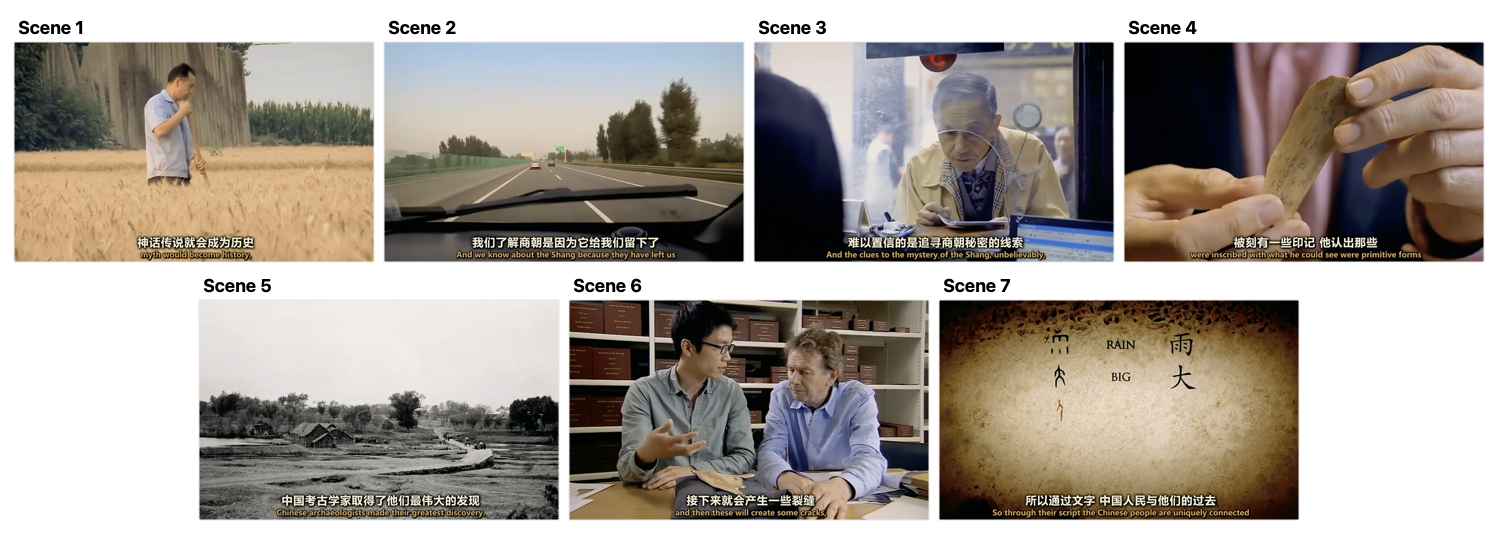}
\end{center}

\begin{caseprompt}
Describe the content of the video by scene and present it in English.
\end{caseprompt}

\begin{caseresponse}
\textbf{Scene 1: Opening Rural Landscape and Mythical History.}
The video begins with a yellow flag waving against a clear sky, framed by lush green trees. The scene transitions to a vast golden wheat field where a farmer harvests crops with a sickle. Children ride bicycles along a rural road, passing by motorbikes. Subtitles state, ``If this was the capital of the Xia, for the Chinese, myth would become history, for they would have found the root of the Chinese state.''

\textbf{Scene 2: Highway Journey and Historical Transition.}
A man drives on a highway, with the rearview mirror reflecting his face. Subtitles mention ``around 1200 BC'' and the search for China's earliest historical rulers, specifically the Shang Dynasty, noted for leaving the ``first Chinese writing.'' The camera shows a roadside statue of a laughing Buddha and highway traffic.

\textbf{Scene 3: Traditional Street and Chinese Medicine Store.}
The scene shifts to a bustling traditional street with pedestrians, a child blowing bubbles, and ornate architectural details. It transitions inside an old Chinese medicine store, where staff prepare herbal packages, and a foreign man navigates the crowded pharmacy. Subtitles highlight the Shang Dynasty's discovery as ``one of the most exciting stories in world archaeology'' and emphasize traditional Chinese medicine's ancient roots.

\textbf{Scene 4: Wang Yirong's Discovery of Oracle Bones.}
Inside the pharmacy, the foreign man recounts the story of Wang Yirong in 1899, a Beijing Hanlin Academy chancellor who, while ill with malaria, purchased medicine containing ``dragon bones,'' namely animal bones. Upon opening the package, he discovered inscriptions resembling ancient bronze writings, leading to the tracing of these bones to Anyang, Henan. Close-ups show him examining the inscribed bones.

\textbf{Scene 5: Anyang Archaeological Excavation.}
Black-and-white historical photos depict the Anyang excavation site, showing massive tombs of late Shang kings with human sacrifices and rows of skulls. Subtitles note that the 1928 excavation uncovered nearly 30,000 oracle bones documenting divination activities of nine Shang kings. The sequence includes images of archaeologists at work and close-ups of inscribed bones.

\textbf{Scene 6: Expert Discussion on Oracle Bone Divination.}
Two experts discuss the oracle bone divination process. One explains how Shang kings burned cracks in turtle shells or ox bones to communicate with ancestors, interpreting crack patterns to determine auspiciousness. They compare oracle bone characters with modern Chinese script, such as the character for ``rain,'' illustrating its evolution from pictographs to modern form.

\textbf{Scene 7: Evolution of Chinese Script Animation.}
An animated sequence illustrates the transformation from prehistoric oracle bone pictographs to modern Chinese script, showing thousands of characters linking the present to the past. Subtitles emphasize how Chinese people are uniquely connected to their history and thinking through their script, surpassing other cultures globally.
\end{caseresponse}

\captionof{figure}{\textbf{Video case for long-form scene-level understanding.} Keye-VL-2.0 summarizes a historical documentary clip by segmenting the video into coherent scenes, tracking transitions from rural landscapes and travel shots to traditional medicine, oracle-bone discovery, Anyang excavation, expert discussion, and the evolution of Chinese script.}
\label{fig:history_video_case}

\clearpage

\subsubsubsection{Case V: Daily Vlog and Equestrian Lesson Understanding}
\label{case_video_ks_daily_vlog}

\begin{center}
\centering
\CaseInputTitle{Input Video}\\[0.4em]
\includegraphics[width=0.66\linewidth]{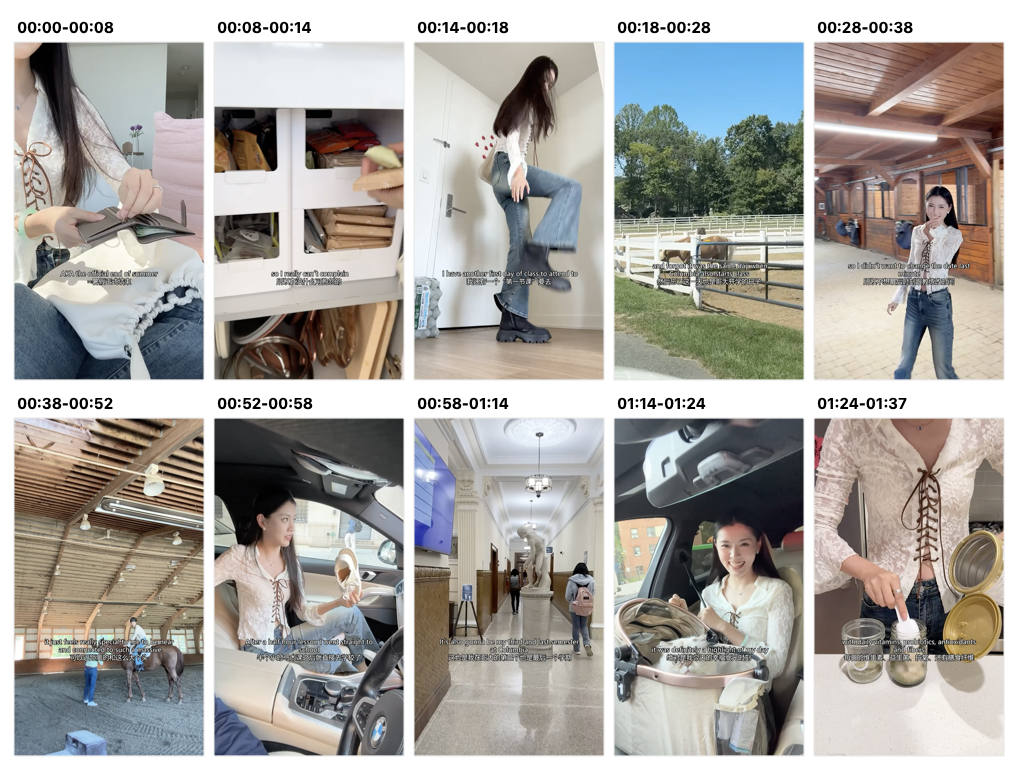}
\end{center}

\begin{caseprompt}
Describe the content of the video by scene and present it in English.
\end{caseprompt}

\begin{caseresponse}
The model segments a daily vlog into ten timestamped scenes, preserving the user's school preparation, equestrian lesson, campus classes, family interaction, and evening routine. The response below is presented as a compact timeline.

\renewcommand{\arraystretch}{1.08}
\begin{tabular}{p{0.18\linewidth}p{0.77\linewidth}}
\toprule
\textbf{Time} & \textbf{Scene summary} \\
\midrule
00:00--00:08 & \textbf{Living room preparation for school.} In a bright living room at 11:39 AM, the woman shows a white drawstring bag and its contents, including a wallet, tablet, and water bottle. On-screen text introduces the first day of class at Columbia and the end of summer break. \\
00:08--00:14 & \textbf{Kitchen snack packing.} She opens organized kitchen drawers, selects snack packets such as black sesame cakes, and packs them for a practical school day. \\
00:14--00:18 & \textbf{Changing shoes for a riding lesson.} She switches into black chunky boots, prepares a white tote bag, and notes that she also has a first riding lesson that day. \\
00:18--00:28 & \textbf{Driving to the stable.} She drives through an automatic gate, sees horses in paddocks, and enters a stable where a farrier is shoeing a horse. \\
00:28--00:38 & \textbf{Riding lesson preparation.} With an instructor's help, she puts on a helmet, mounts the horse, and stands ready for the lesson while expressing affection and trust toward horses. \\
00:38--00:52 & \textbf{Riding lesson.} In an indoor arena, she practices balance and control while the instructor guides the horse around the ring. \\
00:52--00:58 & \textbf{Driving to school.} After the lesson, she changes shoes in the car and buys coffee before class, noting that the timing works out smoothly. \\
00:58--01:14 & \textbf{Campus scenes and classes.} She walks through a historic campus hallway with coffee, attends two classes, and reflects positively on her third and final semester at Columbia. \\
01:14--01:24 & \textbf{Family surprise with cat.} Returning to the car, she finds that her husband has brought their cat Fengfeng in a carrier; she hugs the cat and feels supported by her family. \\
01:24--01:37 & \textbf{Evening health routine.} Back home, she prepares vitamins, probiotics, fiber supplements, and green smoothies with her husband, ending the day with a toast and thanks to the viewer. \\
\bottomrule
\end{tabular}
\renewcommand{\arraystretch}{1}
\end{caseresponse}

\captionof{figure}{\textbf{Video case for scene-level daily vlog understanding.} Keye-VL-2.0 follows a long-form personal vlog across preparation for school, snack packing, a riding lesson, campus classes, a family surprise, and an evening health routine, preserving both temporal boundaries and fine-grained lifestyle details.}
\label{fig:ks_daily_vlog_case}

\subsubsubsection{Case VI: Multi-Domain Service Agent}
\label{agentic_cases}

\begin{caseprompt}
A user asks the assistant to arrange several services for parents visiting Quanzhou on the same day. The request evolves across multiple turns: recommend and book a comfortable hotel near the user's home, choose a spot-removing skincare product that can arrive within one hour, pay the hotel and delivery orders, recommend a Minnan-cuisine restaurant with Ginger Duck and a rich family set meal, create the restaurant order, and finally pay it.
\end{caseprompt}

\begin{caseresponse}
\textbf{Personalized planning.} The assistant first retrieves the user's profile to ground the task in verified information. It identifies the home address, infers that the user's most frequent hotel brand is Home Inn, converts the home address to coordinates, and searches for nearby hotels in Quanzhou. It recommends \textbf{Home Inn (Quanzhou Tian'an Road Branch)} because it is 1.2 km from the user's home, matches the user's historical hotel preference, provides a comfortable queen-bed room, and costs only 288 RMB within the 1000 RMB budget.

\textbf{Hotel booking.} After the user confirms the recommendation, the assistant creates a hotel order for one night in the comfortable queen-bed room:
\begin{center}
\begin{tabular}{ll}
\toprule
\textbf{Item} & \textbf{Result} \\
\midrule
Hotel & Home Inn (Quanzhou Tian'an Road Branch) \\
Room & Comfortable queen-bed room \\
Date & 2024-12-30 \\
Order ID & OOd651df337d \\
Price & 288 RMB \\
\bottomrule
\end{tabular}
\end{center}

\textbf{Delivery recommendation and timing verification.} For the skincare request, the assistant searches for spot-removing skincare products, retrieves store locations, computes distances from the user's home, and converts distances into delivery times. It recommends \textbf{Yuquan Spot-Removing Serum} from \textbf{Yueyan Skincare Store}, because the store is 412 meters away and the estimated delivery time is 28 minutes, satisfying the one-hour constraint.

\textbf{Cross-domain order execution.} The assistant then creates and pays the delivery order, handles the hotel payment with the correct hotel-payment tool after detecting a payment-type mismatch, and searches in-store restaurant options for Minnan cuisine with Ginger Duck. It recommends \textbf{Minnan Flavor Restaurant} and its \textbf{Traditional Minnan Family Banquet Set (3--4 people)} because the set contains Ginger Duck, is suitable for a family meal, and costs 398 RMB. After user confirmation, it creates and pays the in-store order.

\textbf{Final confirmed arrangement.}
\begin{center}
\begin{tabular}{llll}
\toprule
\textbf{Service} & \textbf{Selection} & \textbf{Order ID} & \textbf{Status} \\
\midrule
Hotel & Home Inn queen-bed room, 288 RMB & OOd651df337d & Paid \\
Delivery & Yuquan Spot-Removing Serum, 258 RMB & OTad2c5d4616 & Paid; arrives at 15:58 \\
Restaurant & Minnan family banquet set, 398 RMB & OI3ddaed0869 & Paid \\
\bottomrule
\end{tabular}
\end{center}

\textbf{Agent capability demonstrated.} This case shows that Keye-VL-2.0 can maintain a multi-turn service plan, use user history for personalization, call tools across hotel, delivery, geolocation, distance estimation, and in-store ordering domains, recover from an incorrect payment-tool route, and complete all user-confirmed transactions while keeping the final response concise.
\end{caseresponse}

\captionof{figure}{\textbf{Agent case for multi-domain service orchestration.} Keye-VL-2.0 coordinates personalization, search, geolocation, delivery-time estimation, booking, payment, recommendation, and order creation across hotel, delivery, and in-store service domains.}
\label{fig:multi_domain_service_agent_case}

\clearpage

\section{Contribution (Alphabetical order)}

\medskip
\noindent\textbf{Core Contributors}\par
\bigskip

Bin Wen, Changyi Liu, Chengru Song, Chongling Rao, Guowang Zhang,
Han Li, Haonan Fan, Hengrui Ju, Jiankang Chen, Jiapeng Chen,
Jiawei Yuan, Kaixuan Yang, Kaiyu Jiang, Kun Gai, Lingzhi Zhou,
Na Nie, Sen Na, Tianke Zhang, Tingting Gao, Xuanyu Zheng, Yulong Chen

\bigskip
\noindent\textbf{Contributors}\par
\bigskip

\textbf{Major.  }
Fan Yang, Haixuan Gao, Lele Yang, Mingqiao Liu, Muxi Diao, Qi Zhang,
Qile Su, Wei Chen, Wentao Hong, Xingyu Lu, Yancheng Long,
Yankai Yang, Yingxin Li, Yiyang Fan, Yu Xia, Yuzhe Chen, Ziliang Lai

\medskip
\textbf{Active.  }
Chuan Yi, Haonan Jia, Tianming Liang, Weixin Xu, Xiaoxiao Ma,
Yang Tian, Yufei Han

\bigskip
\noindent\textbf{Supporting Contributors}\par
\bigskip

Feng Han, Hang Li, Jing Wang, Jinghui Jia, Junmin Chen, Junyu Shi,
Ruilin Zhang

\vfill

\begin{center}
    \includegraphics[width=\linewidth]{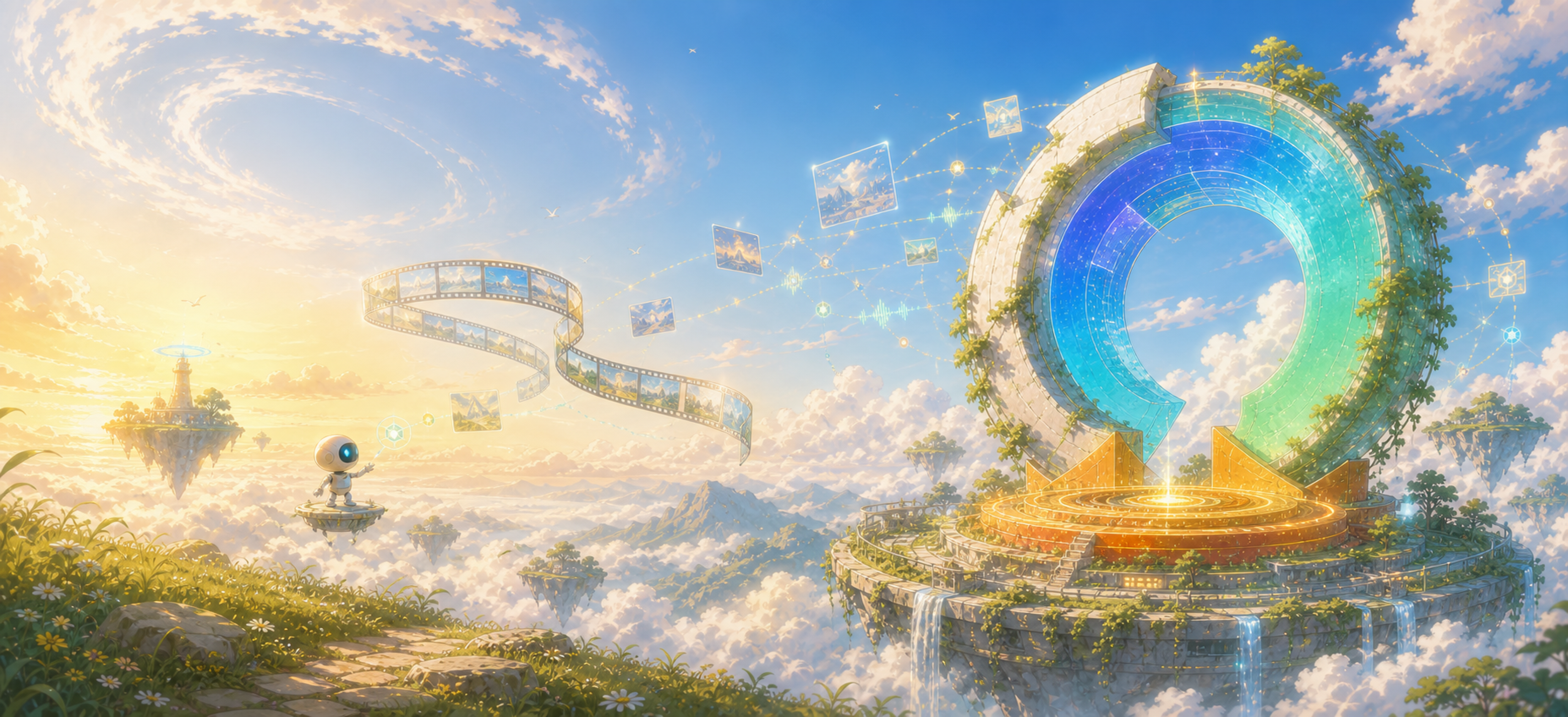}
\end{center}

\end{document}